\documentclass[letterpaper]{article} % DO NOT CHANGE THIS
\usepackage[draft]{aaai2027}  % DO NOT CHANGE THIS
\usepackage[hyphens]{url}  % DO NOT CHANGE THIS
\usepackage{graphicx} % DO NOT CHANGE THIS
\usepackage{natbib}  % DO NOT CHANGE THIS AND DO NOT ADD ANY OPTIONS TO IT
\usepackage{caption} % DO NOT CHANGE THIS AND DO NOT ADD ANY OPTIONS TO IT
\usepackage{algorithm}

\usepackage{algpseudocode}

\usepackage{multirow}
\usepackage{array}
\usepackage{xcolor}
\usepackage{colortbl}
\usepackage{arydshln}
\usepackage{graphicx}
\usepackage{enumitem} 

\usepackage{amsmath}
\usepackage{amssymb}

\usepackage{tabularx}
\usepackage{makecell}
\usepackage{pifont}

\usepackage{subcaption}

\usepackage{threeparttable}
\usepackage[table]{xcolor}

\usepackage{newfloat}
\usepackage{listings}
\DeclareCaptionStyle{ruled}{labelfont=normalfont,labelsep=colon,strut=off} % DO NOT CHANGE THIS
\floatstyle{ruled}
\newfloat{listing}{tb}{lst}{}
\floatname{listing}{Listing}

\usepackage{booktabs}

\title{SciMIF: Understanding Multimodal Instruction Following in
Scientific Domains}
\author{
    Ye Shen\equalcontrib\textsuperscript{\rm 1,2}, Yuting Zheng\equalcontrib\textsuperscript{\rm 1,2}, Dun Pei\textsuperscript{\rm 1}, Zijian Chen\textsuperscript{\rm 1,2}, \\
    Wenlong Zhang\textsuperscript{\rm 1}, Qi Jia\textsuperscript{\rm1},
    Guangtao Zhai\textsuperscript{\rm1,2}
}
\affiliations{
    \textsuperscript{\rm 1}Shanghai Artificial Intelligence Laboratory,
    \textsuperscript{\rm 2}Shanghai Jiao Tong University
}

\begin{document}

\maketitle


\begin{abstract}
Understanding instruction-following capabilities in scientific domains is essential for effectively leveraging Multimodal Large Language Models (MLLMs) to advance the development of scientific fields.  In this work, we introduce SciMIF, a novel benchmark designed to evaluate the capability of MLLMs in following complex scientific instructions. Specifically, based on an extensive analysis of 22 distinct tasks across 5 representative scientific disciplines, we propose a comprehensive taxonomy comprising 10 constraint groups that captures both general functional requirements and discipline-specific characteristics. Guided by this taxonomy, we develop a high-fidelity instruction injection pipeline to systematically augment existing scientific datasets. We conduct comprehensive experiments on multiple state-of-the-art closed-source and open-source MLLMs. Our findings reveal significant performance disparities across different scientific disciplines, with chemistry posing greater challenges for current MLLMs. Furthermore, we observe that increasing the model scale does not yield corresponding improvements in constraint adherence, and current models still struggle severely with fine-grained constraints and instructions requiring the deep application of disciplinary knowledge. SciMIF fills the current void in evaluating multimodal instruction adherence within scientific domains, laying a crucial foundation for future enhancements of MLLMs in rigorous scientific applications. Data and code will be released at \url{https://github.com/shenye7436/SciMIF}.
\end{abstract}

\section{Introduction}
\label{sec:intro}

The application of Multimodal Large Language Models (MLLMs) in scientific domains has rapidly evolved from basic, single-discipline question-answering tasks to complex paradigms such as autonomous science agents and AI-driven scientific discovery~\cite{ai4s-applications,ai4s-chem}. Consequently, the requirements placed on these models have expanded from factual knowledge retrieval to the reliable execution of multi-step scientific actions under explicit operational requirements.

Instruction following is a fundamental capability that transforms models from text-completion engines into versatile task solvers~\cite{instructiongpt,self-instruct}. Following the terminology of CFBench~\cite{csr&isr}, a \textit{constraint} refers to an individual requirement, whereas an \textit{instruction} denotes a complete query containing one or more constraints. In general domains, benchmarks such as IF-Eval~\cite{if-eval} and FollowBench~\cite{followbench} have advanced the evaluation of general-purpose instruction following by systematically assessing whether models satisfy diverse constraints on output format, length, structure, and style. However, it remains underexplored whether current models can reliably follow scientific instructions involving domain-specific constraints.

\begin{figure*}[t!]
    \centering
    \includegraphics[width=\linewidth]{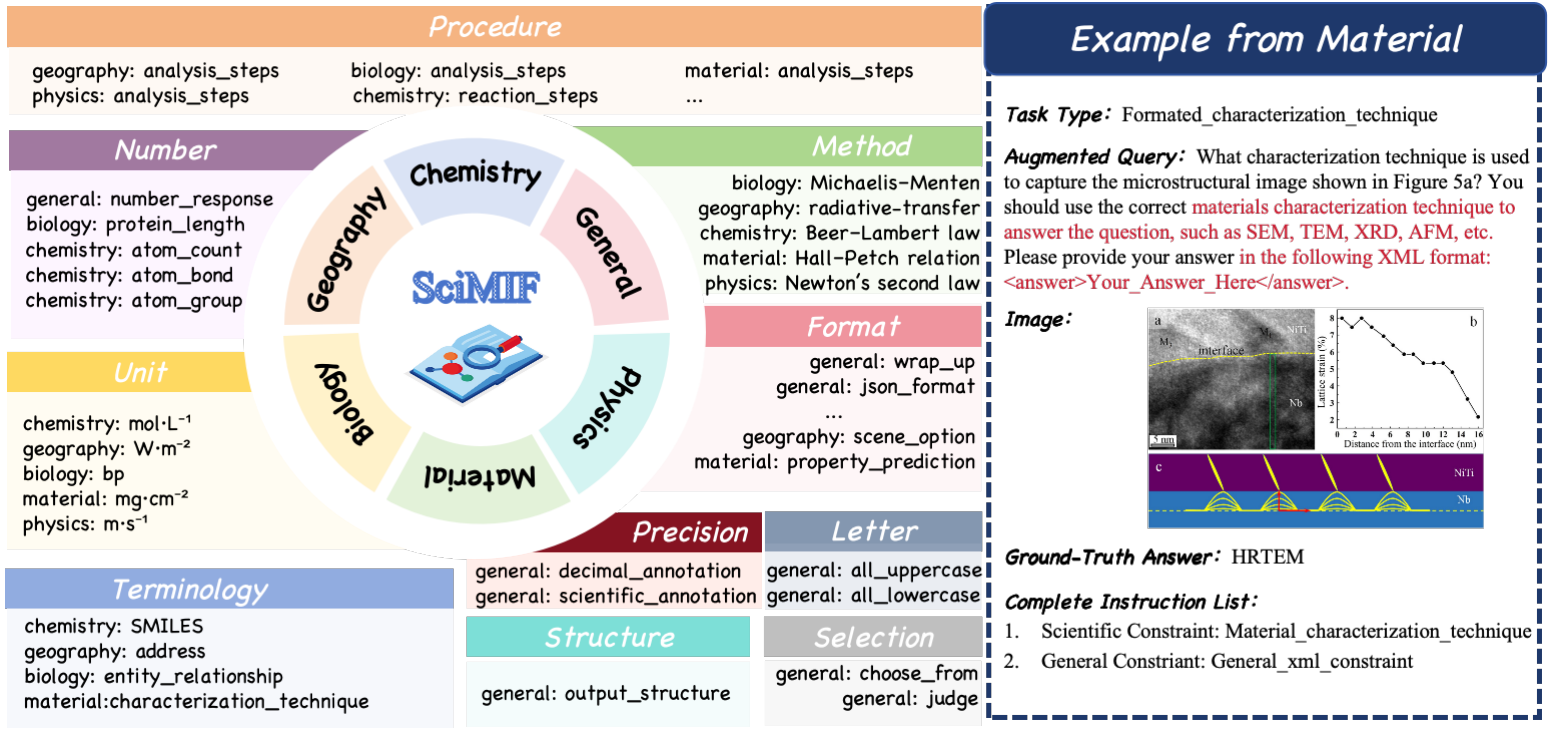}
    \caption{
    An overview of SciMIF, illustrating its coverage of five scientific disciplines, one general constraint domain, and ten functional constraint categories for constructing diverse scientific instruction-following constraints.
    %An overview of SciMIF covering five scientific disciplines, a general constraint domain, and ten functional constraint groups. 
    }
    \label{fig:constraint}
\end{figure*}

Scientific instruction following differs from its general-domain counterpart in three important aspects. First, scientific instructions are \textbf{scientific knowledge-dependent}. Even seemingly simple constraints involving letters, numbers, terminology, or output formats may carry domain-specific meanings that cannot be handled through surface-level text manipulation alone. For example, constraints involving the numbers of chemical bonds, functional groups, or amino-acid residues require models to accurately recognize and understand the relevant scientific entities.
 Second, the semantics of scientific instructions exhibit strong \textbf{interdisciplinary variations}.  Although different disciplines may share functional constraint groups, their concrete meanings differ substantially across scientific contexts. As illustrated in Figure~\ref{fig:constraint}, a terminology constraint may require valid molecular nomenclature in chemistry, a hierarchical address in geography, an entity relationship in biology, or a characterization technique in materials science.  Third, scientific instruction following is frequently \textbf{multimodal}. Scientific reasoning often depends on specialized visual inputs, including molecular structures, biological diagrams, materials microscopy images, geographical scenes, and physical plots. For instance, in the materials example in Figure~\ref{fig:constraint}, identifying the required characterization technique depends jointly on the textual request and the information presented in the scientific figure.

These characteristics also expose a practical limitation of existing scientific evaluations. Most scientific benchmarks primarily determine whether a final answer is correct, but scientific correctness alone does not guarantee that an output is usable. A scientifically correct answer may still violate a requested unit, notation, output structure, analysis method, or solution procedure. Conversely, a model may strictly reproduce the requested form while relying on scientifically invalid reasoning. Treating these outputs as a single notion of correctness obscures two distinct model capabilities. We therefore distinguish \textit{scientific correctness}, which measures whether an answer is scientifically valid, from \textit{instruction adherence}, which measures whether the specified constraints are satisfied. Evaluating them separately enables failures to be attributed to insufficient scientific reasoning, inadequate instruction following, or both.

To address these challenges, we introduce SciMIF, a \textbf{Sci}entific \textbf{M}ultimodal \textbf{I}nstruction \textbf{F}ollowing benchmark for systematically evaluating models across diverse scientific questions. As illustrated in Figure~\ref{fig:constraint}, SciMIF covers five representative disciplines: chemistry, geography, biology, materials science, and physics. We tasked domain experts with systematically examining 22 scientific tasks, their source datasets, disciplinary conventions, and related literature, and deriving a taxonomy of ten functional constraint groups. These groups capture shared instruction-following capabilities across disciplines, while their concrete constraints are adapted to domain-specific knowledge and practices.

Based on this taxonomy, we develop a scalable framework for transforming existing scientific problems into instruction-following evaluations. The framework first identifies constraints already implicit in the original tasks and separates them into independently evaluable requirements. It then injects additional compatible scientific and general constraints without changing the correct. After automatic consistency checking and human verification, the benchmark contains 2,527 samples and supports evaluation evaluation of constraint satisfaction, complete instruction fulfillment, and fine-grained adherence to decomposed requirement.%criteria.
%at the constraint, instruction, and decomposed-requirement levels.

Extensive experiments on representative closed-source and open-source MLLMs reveal substantial limitations in scientific instruction following. Performance varies considerably across disciplines, with chemistry and geography posing greater challenges than other domains. Increasing model scale does not consistently improve constraint adherence, indicating that parameter scaling alone is insufficient. Models also perform substantially worse on general constraints and fine-grained requirements, while scientific correctness and instruction adherence remain weakly coupled. These findings motivate future research on domain-aware instruction alignment, stronger adherence to general operational requirements, and structure-aware or tool-assisted methods for handling fine-grained scientific constraints.

In summary, our main contributions are as follows:
\begin{itemize}
    \item We establish an expert-derived taxonomy of scientific constraints spanning five disciplines and ten functional groups, capturing both shared capability dimensions and discipline-specific requirements.

    \item We propose a scalable framework for converting existing scientific tasks into instruction-following evaluations by recognizing implicit constraints and injecting compatible scientific and general constraints without changing the reference answers.

    \item We construct SciMIF and evaluate representative MLLMs, revealing substantial disciplinary variation, difficulties with fine-grained constraints, and a clear gap between scientific correctness and instruction adherence.
\end{itemize}
\section{Related Work}
\label{sec:related_work}

\subsection{Scientific Reasoning Benchmarks}

Scientific benchmarks have evolved from general science question answering toward more specialized and challenging reasoning tasks. ScienceQA~\cite{saikh2022scienceqa} provides elementary and secondary school science questions covering multiple disciplines. Subsequently, MMMU~\cite{yue2024mmmu} has pushed the difficulty to the expert level by introducing university-level expertise. The core of scientific tasks lies in the construction of logical chains. MathVista~\cite{lu2023mathvista} is designed specifically for multimodal mathematical reasoning, emphasizing the model's ability to handle complex visual elements such as geometric figures and function graphs. SciBench~\cite{wang2023scibench} further focuses on complex scientific computing problems, challenging the limits of the model's multi-step reasoning and formula calculation through in-depth coverage of disciplines such as physics and chemistry. In addition, in-depth evaluations of specific fields are also crucial. For example, in the field of geography, GeoQA~\cite{chen2021geoqa} examines the model’s perception of spatial layout and geographical features. Although these benchmarks have substantially advanced the evaluation of scientific knowledge and reasoning, they primarily measure whether models produce correct answers, without systematically examining whether the outputs satisfy explicit scientific and operational constraints.

% Required packages
% \usepackage{booktabs}
% \usepackage{tabularx}
% \usepackage{array}
% \usepackage{makecell}
% \usepackage[table]{xcolor}
% \usepackage{pifont}

\newcommand{\cmark}{\ding{51}}
\newcommand{\xmark}{\ding{55}}

% Smaller citation font for compact tables
\newcommand{\tabcite}[1]{%
    \nobreak\hspace{0.15em}%
    {\fontsize{5.8}{6.2}\selectfont\cite{#1}}%
}

\begin{table}[t]
    \centering
    \fontsize{7.3}{8.2}\selectfont
    \renewcommand{\arraystretch}{1.08}
    \setlength{\tabcolsep}{0.7pt}

    \begin{tabularx}{\columnwidth}{
        @{}
        >{\raggedright\arraybackslash}X
        >{\centering\arraybackslash}p{0.75cm}
        >{\centering\arraybackslash}p{0.80cm}
        >{\centering\arraybackslash}p{0.90cm}
        >{\centering\arraybackslash}p{0.80cm}
        >{\raggedleft\arraybackslash}p{0.95cm}
        >{\centering\arraybackslash}p{0.80cm}
        @{}
    }
        \toprule
        \textbf{Benchmark}
        & \textbf{Scope}
        & \makecell{\textbf{\# Grp.}}
        & \makecell{\textbf{\# Const.}}
        & \makecell{\textbf{\# Disc.}}
        & \textbf{\# Samp.}
        & \makecell{\textbf{Multi-}\\\textbf{modal}} \\
        \midrule

        \multicolumn{7}{@{}l}{
            \textit{General-purpose instruction-following benchmarks}
        } \\[0.7pt]

        IFEval\tabcite{if-eval}
        & General
        & 9
        & 25
        & --
        & 541
        & \xmark \\
        
        FollowBench\tabcite{followbench}
        & General
        & 5
        & 16
        & --
        & 820
        & \xmark \\

        MM-IFEngine\tabcite{mm-ifengine}
        & General
        & 6
        & 32
        & --
        & 400
        & \cmark \\

        \addlinespace[1pt]
        \midrule

        \multicolumn{7}{@{}l}{
            \textit{Scientific instruction-following benchmarks}
        } \\[0.7pt]

        % LLaSMol\tabcite{llasmol}
        % & Scientific
        % & 4
        % & 14
        % & 1
        % & 3.3M
        % & \xmark \\

        SciIF\tabcite{sciif}
        & Scientific
        & 3
        & 10
        & 4
        & 1,244
        & \xmark \\

        \rowcolor{gray!10}
        \textbf{SciMIF (Ours)}
        & \textbf{Both}
        & \textbf{10}
        & \textbf{42}
        & \textbf{5}
        & \textbf{2,527}
        & \cmark \\

        \bottomrule
    \end{tabularx}
    \caption{
        Comparison of general-purpose and scientific instruction-following
        benchmarks.
        \textit{Scope} specifies whether they evaluate general, scientific,
        or both types of constraints. \# Grp. and \# Const. denote the numbers
        of functional constraint groups and individual constraints; \# Disc.
        and \# Samp. denote covered scientific disciplines and evaluation
        samples. ``--'' indicates scientific discipline coverage is not applicable.
    }
    \label{tab:benchmark_comparison}
\end{table}

\subsection{Instruction-Following Benchmarks}

General-purpose instruction-following benchmarks evaluate whether models comply with explicit user requirements. IFEval~\cite{if-eval} introduces objectively verifiable textual constraints, FollowBench~\cite{followbench} evaluates constraints of varying difficulty and their combinations, and MM-IFEngine~\cite{mm-ifengine} extends instruction-following evaluation to multimodal inputs. However, these benchmarks mainly focus on general linguistic, formatting, and structural requirements, without covering constraints grounded in scientific knowledge and disciplinary conventions.
Scientific instruction-following evaluation remains limited.
% LLaSMol~\cite{llasmol} focuses on chemistry-related instructions, while 
SciIF~\cite{sciif} evaluates university-level scientific question answering under process-level guidance, such as prescribed reasoning steps and solution procedures. Its constraints are shared across scientific questions rather than systematically derived from the representations, conventions, and operational requirements of individual disciplines, and its evaluation remains text-only. As summarized in Table~\ref{tab:benchmark_comparison}, SciMIF complements existing benchmarks with an expert-derived taxonomy constructed from 22 tasks across five disciplines. It covers 10 functional groups and 42 discipline-adapted constraints, combines scientific and general requirements, and supports multimodal scientific inputs.

\section{SciMIF}
\label{sec:approach}
This section introduces the design of SciMIF including its taxonomy of scientific constraints, instruction-augmented sample construction procedure, and benchmark statistics.

\subsection{Taxonomy of Scientific Constraints}
\label{sec:constraint_categories}

% To support fine-grained evaluation of scientifically grounded requirements, we construct a hierarchical taxonomy of scientific constraints. Rather than treating these constraints as generic linguistic templates, we derive the taxonomy through an expert-guided analysis of 22 tasks from 13 scientific datasets. Domain experts examine the original task definitions, queries, reference answers, disciplinary conventions, evaluation protocols, and relevant literature to identify recurring operational requirements. Requirements serving similar functional purposes are grouped into shared categories, while their discipline-specific meanings are retained as individual constraints.

% To provide a comprehensive evaluation of scientific instruction following, we construct a hierarchical taxonomy covering both general and discipline-specific constraints. Although seemingly general groups such as format, structure, and selection are not unique to science, they are integral to scientific instructions and often interact with domain-specific requirements. Considering them therefore enables a more complete assessment of a model's ability to execute scientific instructions. We derive the taxonomy through an expert-guided analysis of 22 tasks from 13 scientific datasets. Domain experts examine the original tasks, reference answers, disciplinary conventions, evaluation protocols, and relevant literature to identify recurring requirements, which are grouped by functional purpose while retaining their discipline-specific meanings as individual constraints.

To provide a comprehensive evaluation of scientific instruction following, we construct a hierarchical taxonomy covering both general and discipline-specific constraints. In real-world scientific queries, user requirements and agent interfaces often combine general operational constraints with scientific requirements, making the two inseparable in practical use. Our taxonomy therefore considers both: general constraint types are adapted from prior instruction-following benchmarks, while the primary effort focuses on constructing discipline-specific constraints. Specifically, domain experts analyze 22 tasks from 13 scientific datasets, together with their reference answers, disciplinary conventions, evaluation protocols, and relevant literature, to identify recurring scientific requirements. Functionally similar requirements are abstracted into shared constraint groups, while their domain-specific realizations are preserved as individual constraints.

This results in a two-level organization: the domain level identifies the field of the corresponding instruction, covering one general domain and five scientific domains, whereas the group level identifies the functional purpose of each constraint. Each individual constraint is therefore characterized by both a domain and a group, while each functional group may contain distinct domain-specific constraints, as shown in Figure~\ref{fig:constraint}. This organization enables model performance to be analyzed from both disciplinary and capability-oriented perspectives. Detailed domain-specific instantiations and complete constraint specifications are provided in the Appendix~\ref{app:constraints}.

At the group level, we define ten functional types according to the requirements imposed on model outputs and task execution:
\begin{itemize}[leftmargin=0.5cm, noitemsep, topsep=0pt]
    \item \textbf{Procedure:} Requires reasoning, analysis, or experimental operations to follow a prescribed sequence of steps.

    \item \textbf{Number:} Specifies the required cardinality of output items or scientific entities, such as responses, atoms, bonds, functional groups, or sequence elements.

    \item \textbf{Method:} Requires the use of a designated scientific formula, theorem, law, or analytical approach.

    \item \textbf{Unit:} Requires numerical quantities to be expressed in specified units while preserving dimensional consistency.

    \item \textbf{Format:} Specifies the required representation of scientific outputs, such as expressing material properties as discrete labels or continuous values, or organizing geographic locations in a prescribed hierarchical format.

    \item \textbf{Terminology:} Requires valid and standardized nomenclature or specialized expressions appropriate to a scientific discipline.

    \item \textbf{Precision:} Requires numerical quantities to be reported with a specified level of accuracy, such as a fixed number of decimal places or significant figures.

    \item \textbf{Letter:} Specifies general letter-level requirements, such as using only lowercase or uppercase letters.

    \item \textbf{Structure:} Specifies how different parts of the response should be organized, such as presenting the analysis before the final answer.

    \item \textbf{Selection:} Restricts the answer to one or more valid choices from a predefined candidate set.
\end{itemize}

Although these functional groups are shared across disciplines, their concrete instantiations may require substantially different scientific knowledge and recognition capabilities. For example, a terminology constraint may involve valid molecular nomenclature in chemistry, a hierarchical address in geography, an entity relationship in biology, or an appropriate characterization technique in materials science. Similarly, a number constraint may refer to atoms and chemical bonds in chemistry, sequence elements in biology, or physical quantities in physics, while a procedure constraint may specify a reaction sequence, a biological analysis pipeline, an experimental workflow, or a physics derivation. These discipline-specific instantiations allow SciMIF to analyze model performance both across scientific disciplines and along shared constraint-following capabilities.

\begin{figure}
    \centering
    \includegraphics[width=0.98\linewidth]{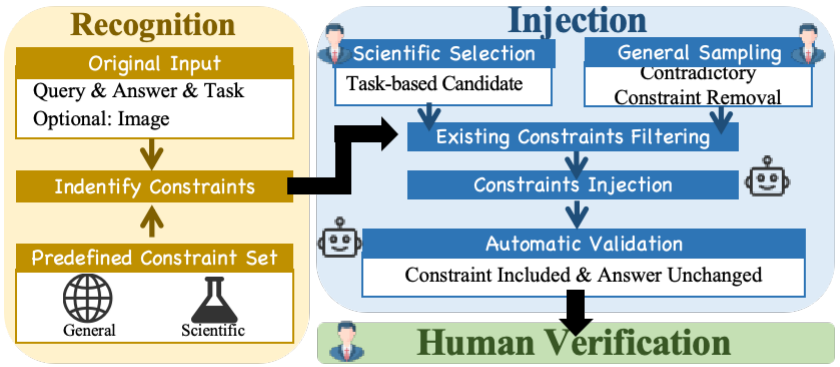}
    \caption{An overview of the data construction pipeline.}
    \label{fig:approach}
\end{figure}

\subsection{Data Construction}
\label{sec:data_construction}
As illustrated in Figure~\ref{fig:approach}, we construct SciMIF through a structured pipeline consisting of seed preparation, constraint recognition and selection, constraint injection, and human verification. We provide more details of the construction pipeline in the Appendix~\ref{app:construction_details}.
\paragraph{Seed Preparation.}
Each seed sample from a source dataset contains a textual question $q$, an optional visual input $I$, a reference answer $a$, and a scientific task type $t$. We represent the sample as
\begin{equation}
    x = (q, I, a, t).
\end{equation}
For each sample, we determine an applicable constraint inventory $C_x$, which contains scientific constraints compatible with its discipline and task type, together with general constraints applicable to scientific tasks. 

\paragraph{Constraint Recognition.}
Annotators first identify the constraints already expressed or implied by the original query, which can be expressed as a set $C_o$:
\begin{equation}
    C_o = f_r(q, C_x),
\end{equation} 

The remaining constraints in $C_x \setminus C_o$ are then filtered to remove requirements that are redundant, incompatible with the task, or contradictory to existing constraints. Scientific constraints $C_{s}$ are selected according to the disciplinary knowledge and task type, while mutually compatible general constraints $C_{g}$ are randomly sampled by $N$ classes from $C_x \setminus C_o$.

\paragraph{Constraint Injection.}
%The selected scientific constraints are first adapted to the original query. We then sequentially inject up to $N$ compatible general constraints. 
Scientific constraints from $C_{s}$ should be injected to original query $q$ as follows:
\begin{equation}
    q_d=f_d(q, C_{s}),
\end{equation}
where $q_d$ represents the augmented query after injecting appropriate scientific constraints. 

For general constraints in $C_{g}$, if one constraint is injected into $q_d$, the new augmented query can be expressed as:
\begin{equation}
    q_g = f_g(q_d, c_n\mid c_n\subseteq C_{g}),
\end{equation}
where $c_n$ denotes any one of each general constraint class. $f_g (\cdot)$ denotes a double automatic validation procedure consisting of $\mathbb{I}_{included}(q_g, c_n) \cap \mathbb{I}_{unchanged}(q_g, a)$, which ensures that $c_n$ has been included and ground-truth answer $a$ remains unchanged. A failed injection is retried up to $k$ times using an alternative constraint from the same category. 

The resulting sample after the validation is represented as
\begin{equation}
    \hat{x} = (\hat{q}, I, a, t),
\end{equation}
where $\hat{q}$ denotes the final augmented query. 

\begin{table}[t]
\centering
\small
\setlength{\tabcolsep}{3pt}

\begin{tabularx}{\columnwidth}{
    @{}
    >{\raggedright\arraybackslash}X
    c c c c c
    @{}
}
\toprule
Statistic & Chem. & Geo. & Bio. & Mat. & Phy. \\
\midrule
Total                  & 518 & 523 & 493 & 497 & 496 \\
Scientific Constraints & 7 & 5 & 5 & 5 & 3 \\
Task Types             & 8 & 4 & 3 & 4 & 3 \\
Avg. Question (/tokens) & 231.46 & 168.17 & 147.29 & 215.60 & 293.68 \\
Multimodal Samples     & $\times$ & $\checkmark$ & $\times$
                       & $\checkmark$ & $\checkmark$ \\
\bottomrule
\end{tabularx}
\caption{Overall statistics of the constructed dataset.}
\label{tab:data_stats_new}
\end{table}

\paragraph{Human Verification.}
Two annotators ask for each $\hat{x}$ and its associated constraint list meet the following two requirements: (1) \textit{Logical Coherence and Fluency}, ensuring that the injected constraints are naturally integrated without semantic contradictions or grammatical errors; and (2) \textit{Constraint Fidelity}, ensuring that every specified constraint is accurately and unambiguously expressed in the query. Problematic samples are manually revised, while samples that cannot be meaningfully repaired are discarded. All retained samples undergo manual verification, and 884 samples are revised during this process. 

{

\begin{table*}[!t]
\centering
\small
\setlength{\tabcolsep}{0.5pt}
 
% \begin{adjustbox}{width=\linewidth}
% \renewcommand{\arraystretch}{1.2}
\begin{tabular*}{\textwidth}{@{\extracolsep{\fill}}l|*{5}{ccc|}c@{}}
\hline

\multirow{2}{*}{\textbf{Model}} 
& \multicolumn{3}{c|}{\textbf{Chemistry}} 
& \multicolumn{3}{c|}{\textbf{Geography}} 
& \multicolumn{3}{c|}{\textbf{Biology}} 
& \multicolumn{3}{c|}{\textbf{Material}} 
& \multicolumn{3}{c|}{\textbf{Physics}} 
& \multirow{2}{*}{\textbf{Overall}} 
\\
\cline{2-4} \cline{5-7} \cline{8-10} \cline{11-13} \cline{14-16}

& \textbf{CSR} 
& \textbf{ISR} 
& \textbf{DRFR} 
& \textbf{CSR} 
& \textbf{ISR} 
& \textbf{DRFR} 
& \textbf{CSR} 
& \textbf{ISR} 
& \textbf{DRFR} 
& \textbf{CSR} 
& \textbf{ISR} 
& \textbf{DRFR} 
& \textbf{CSR} 
& \textbf{ISR} 
& \textbf{DRFR} 
\\
\hline

% ===== 数据行示例=====
\multicolumn{17}{c}{\textit{Closed-Source MLLMs}}\\
\hdashline
%GPT-5.2 & \textbf{73.09} & \textbf{45.56} & \textbf{74.24} & 82.06 & 60.99 & 82.03 & \textbf{90.59} & \textbf{76.67} & \textbf{73.87} & 75.22 & 43.26 & 73.22 & \textbf{68.22} & \textbf{27.42} & \textbf{64.58} & \textbf{32.99}\\
%GPT-5.2 & \textbf{73.09} & \textbf{45.56} & \textbf{74.24} & 82.06 & \textbf{60.99} & 82.03 & \textbf{90.59} & \textbf{76.67} & \textbf{88.59} & 88.28 & \textbf{78.87} & \textbf{89.21} & \textbf{87.77} & \textbf{68.75} & \textbf{86.65} & \textbf{66.17}\\
\mbox{GPT-5.2} & \textbf{72.87} & \textbf{46.33} & \textbf{73.99} & 82.30 & \textbf{61.57} & 82.40 & \textbf{88.87} & \textbf{72.82} & \textbf{86.58} & 87.79 & \textbf{78.07} & \textbf{88.93} & \textbf{87.97} & \textbf{69.56} & \textbf{86.94} & \textbf{65.67}\\
%Grok-4-Fast & 53.55 & 18.34 & 53.51 & \textbf{62.36} & \textbf{32.70} & \textbf{62.00} & 65.58  & 40.24 & 63.03 & \textbf{77.84} & 49.09 & \textbf{74.11} & 63.81 & 22.58 & 60.09 & 32.59\\
%Grok-4-Fast & 65.56 & 32.82 & 67.17 & \textbf{83.76} & 52.96 & \textbf{82.88} & 80.16  & 45.12 & 78.30 & \textbf{90.10} & 75.05 & 89.20 & 81.74 & 57.46 & 80.72 & 52.68\\
\mbox{Grok-4-Fast} & 65.56 & 33.01 & 67.37 & \textbf{83.96} & 53.15 & \textbf{83.25} & 80.18  & 45.53 & 79.06 & \textbf{89.74} & 75.25 & 89.01 & 82.06 & 58.27 & 81.16 & 53.04\\
%Gemini-3.1-Pro-Preview & 54.18 & 19.31 & 53.27 & 56.53 & 19.69 & 57.38 & 62.97 & 39.55 & 59.14 &77.66 & \textbf{50.50} & 73.70 & 58.16 & 16.13 & 55.95 & 29.04\\
%Gemini-3.1-Pro-Preview & 66.12 & 31.85 & 66.78 & 79.30 & 47.23 & 79.64 & 76.33 & 46.45 & 73.07 &90.10 & 75.65 & 88.95 & 77.44 & 50.20 & 77.79 & 50.26\\
\mbox{Gemini-3.1-Pro-Preview} & 65.79 & 32.24 & 66.37 & 79.53 & 47.80 & 80.01 & 75.30 & 44.02 & 72.01 &89.94 & 75.45 & 89.00 & 77.46 & 50.60 & 77.78 & 50.02\\
%Claude-Sonnet-4.6 & 49.60 & 13.35 & 49.26 & 48.07 & 8.90 & 49.93 & 61.99 & 43.16 & 58.10 & 68.32 & 34.41 & 67.71 & 66.43 & 22.58 & 63.05 & 24.48\\
%Claude-Sonnet-4.6 & 61.84 & 26.31 & 63.07 & 71.01 & 31.33 & 72.40 & 75.83 & 43.79 & 72.56 & 81.17 & 66.40 & 83.53 & 86.15 & 63.91 & 85.42 & 46.35\\
\mbox{Claude-Sonnet-4.6} & 62.02 & 26.31 & 63.42 & 71.31 & 31.72 & 72.75 & 76.56 & 46.32 & 73.75 & 81.54 & 67.40 & 84.06 & 86.48 & 65.12 & 85.87 & 47.37\\
\hline
\multicolumn{17}{c}{\textit{Open-Source MLLMs}}\\
\hdashline
%Qwen3.5-27B & 56.16 & 18.31 & 56.26 & 56.50 & 20.00 & 57.37 & 66.32 & 45.26 & 62.50 & 76.41 & 44.24 & 73.25 & 65.54 & 19.60 & 62.68 & 29.48\\
%Qwen3.5-27B & 68.54 & 31.89 & 70.21 & 79.21 & 44.23 & 79.50 & 84.19 & 51.58 & 81.77 & 89.32 & \textbf{68.52} & 89.06 & 85.30 & 60.61 & 85.10 & 51.37\\
\mbox{Qwen3.5-27B} & 68.33 & 32.09 & 70.04 & 79.54 & 45.00 & 80.03 & 83.18 & 49.21 & 80.79 & 89.76 & \textbf{69.75} & 89.62 & 85.74 & 60.82 & 85.70 & 51.37\\
%Qwen3.5-35B-A3B & 56.48 & 18.42 & 56.71 & 54.28 & 17.95 & 55.41 & 68.56 & 47.11 & 63.84 & 76.74 & 44.35 & 73.47 & 64.91 & 17.54 & 62.36 & 29.07\\
%Qwen3.5-35B-A3B & 68.87 & 30.10 & 70.77 & 77.23 & 40.73 & 77.78 & 85.10 & 51.98 & 82.05 & 90.02 & 68.41 & 89.66 & 84.70 & 59.48 & 84.81 & 50.14\\
\mbox{Qwen3.5-35B-A3B} & 68.52 & 29.90 & 70.35 & 77.61 & 41.31 & 78.39 & 83.91 & 49.24 & 81.22 & 90.08 & 68.83 & 89.88 & 85.00 & 60.48 & 85.25 & 49.95\\
%Qwen3.5-122B-A10B & 56.18 & 18.41 & 56.19 & 55.45 & 19.62 & 56.59 & 66.40 & 43.43 & 63.81 & 76.78 & \textbf{45.53} & 73.47 & 65.79 & 19.76 & 63.00 & 29.35\\
%Qwen3.5-122B-A10B & 68.34 & 30.23 & 69.98 & 78.13 & 42.69 & 78.72 & 82.82 & 48.59 & 81.18 & \textbf{89.64} & 68.09 & 89.20 & 85.50 & 61.49 & \textbf{85.38} & 50.22\\
\mbox{Qwen3.5-122B-A10B} & 68.08 & 30.62 & 69.65 & 78.40 & 43.27 & 79.16 & 82.13 & 47.18 & 80.67 & \textbf{89.77} & 68.70 & 89.50 & 85.89 & 62.30 & \textbf{85.90} & 50.41\\
%Qwen3.5-397B-A17B & \textbf{60.38} & \textbf{22.78} & \textbf{59.94} & \textbf{58.79} & \textbf{23.85} & \textbf{59.72} & \textbf{71.94} & \textbf{47.92} & \textbf{69.16} & \textbf{76.99} & 44.31 & \textbf{74.21} & 65.90 & 19.15 & 63.00 & \textbf{31.60} \\
%Qwen3.5-397B-A17B & \textbf{72.29} & \textbf{39.96} & \textbf{73.50} & \textbf{81.27} & \textbf{53.46} & \textbf{81.52} & \textbf{86.68} & \textbf{59.79} & \textbf{84.79} & 89.73 & 73.17 & \textbf{89.81} & 85.55 & 61.49 & 85.30 & \textbf{57.57} \\
\mbox{Qwen3.5-397B-A17B} & \textbf{72.19} & \textbf{39.96} & \textbf{73.49} & \textbf{81.57} & \textbf{54.23} & \textbf{81.97} & \textbf{85.41} & \textbf{56.87} & \textbf{83.52} & 89.63 & 73.37 & \textbf{89.86} & 85.80 & 62.50 & 85.67 & \textbf{57.39} \\
%InternVL3.5-8B & 51.23 & 13.90 & 50.35 & 53.87 & 11.09 & 54.82 & 70.20 & 45.44 & 68.07 & 67.15 & 32.19 & 65.27 & 65.17 & 25.00 & 60.96 & 25.52\\
%InternVL3.5-8B & 62.64 & 31.08 & 63.25 & 72.47 & 44.17 & 72.98 & 78.94 & 56.19 & 77.42 & 79.54 & 60.16 & 80.53 & 84.38 & 62.90 & 82.65 & 50.90\\
\mbox{InternVL3.5-8B} & 62.99 & 31.47 & 63.83 & 73.04 & 44.93 & 73.82 & 79.14 & 57.20 & 77.97 & 79.34 & 59.76 & 80.56 & 84.70 & 63.51 & 83.09 & 51.37\\
%InternVL3.5-14B & 50.18 & 11.58 & 50.55 & 56.53 & 16.63 & 57.32 & 70.59 & 45.64 & 68.25 & 66.89 & 30.78 & 65.98 & \textbf{66.93} & \textbf{27.22} & \textbf{63.05} & 26.37\\
%InternVL3.5-14B & 61.77 & 30.31 & 63.76 & 74.81 & 49.33 & 75.07 & 81.02 & 59.23 & 79.29 & 79.15 & 59.15 & 81.07 & \textbf{86.09} & \textbf{65.93} & 84.75 & 52.79\\
\mbox{InternVL3.5-14B} & 61.88 & 30.89 & 63.96 & 75.23 & 49.90 & 75.75 & 80.48 & 58.62 & 79.05 & 79.33 & 59.76 & 81.43 & \textbf{86.42} & \textbf{66.53} & 85.19 & 53.14\\
%InternVL3.5-38B & 47.85 & 11.20 & 48.58 & 55.31 & 13.96 & 56.07 & 66.94 & 40.97 & 64.68 & 66.13 & 30.99 & 65.25 & 64.89 & 23.79 & 60.92 & 24.18\\
%InternVL3.5-38B & 59.70 & 29.15 & 61.93 & 74.38 & 46.85 & 74.72 & 76.53 & 54.56 & 74.73 & 78.64 & 59.76 & 80.66 & 84.10 & 61.69 & 82.70 & 50.40\\
\mbox{InternVL3.5-38B} & 59.96 & 29.73 & 62.36 & 74.85 & 47.80 & 75.48 & 77.32 & 56.80 & 76.18 & 79.26 & 61.17 & 81.35 & 84.52 & 62.70 & 83.28 & 51.64\\
% DeepSeek-V3.2-Speciale & 38.35 & 11.00 & 35.62 & 38.17 & 5.74 & 39.26 & 50.29 & 31.03 & 47.80 & 57.36 & 21.33 & 52.56 & 37.61 & 4.44 & 38.54 & 14.71 \\
\hline
\end{tabular*}
% \end{adjustbox}
\caption{Performance (\%) of evaluated MLLMs. \textit{Overall} denotes the average ISR across the five scientific disciplines. The best score in each column is bolded separately for closed-source and open-source MLLMs. }
\label{tab:main results}
\end{table*}
}

%\textbf{Declarative Memory} represents the average scores across explicit memory tasks, \textbf{Non-declarative} represents average scores across implicit memory tasks, and \textbf{Overall} represents the average performance of models between declarative memory and non-
\subsection{Data Statistics}
\paragraph{Data Sources.}
SciMIF is constructed from 13 existing scientific datasets, covering 22 task types across five disciplines: chemistry, geography, biology, materials science, and physics. Representative sources include ChemEval~\cite{chemeval}, IMAGEO-Bench~\cite{IMAGEO-Bench}, LAB-Bench~\cite{lab-bench}, MatCha~\cite{matcha}, and PhysUniBench~\cite{phyunibench}. These datasets provide diverse task formulations, disciplinary conventions, and input modalities. The complete source-to-task mapping and task-level sample
statistics are provided in the Appendix~\ref{app:data_sources}.

% \noindent\textbf{Data Sources.} Our dataset is constructed from a diverse array of task sources spanning multiple scientific domains, which map to specific discipline-specific instruction categories, including Chemistry ( S2-TOMG-Bench-mini~\cite{TOMG-Bench}, ChemEval~\cite{chemeval}), Geography ( EarthSE~\cite{earthse}, IMAGEO-Bench~\cite{IMAGEO-Bench}), Biology ( Mol-Instructions~\cite{Mol-Instructions}, Lab-Bench~\cite{lab-bench}), Material ( Mascqa~\cite{mascqa}, MatCha~\cite{matcha}, LLM4Mat-Bench~\cite{llm4mat}, MatSciBench~\cite{matscibench}), and Physics ( UGPhysics~\cite{ugphysics}, Physreason~\cite{physreason}, PhysUniBench~\cite{phyunibench}). Overall, the dataset contains 22 task types collected from 13 sources. This broad and heterogeneous coverage across various scientific disciplines guarantees a highly representative sample, effectively mitigating domain-specific biases and ensuring that the tasks evaluated in our study are structurally diverse and inherently unbiased. The complete list of source datasets and their disciplinary coverage is provided in Appendix.

%用到的数据集+引用

%In the main text, we provide a high-level overview of the dataset construction process, while the detailed mapping between tasks and domain-specific instructions is provided in Appendix~\ref{app:instruction_mapping}. %Specifically, Appendix~\ref{app:instruction_mapping} lists how each task type corresponds to its associated domain constraints and instruction templates.

%Overall, the dataset contains 21 task types collected from xxx sources, covering 5 domains and 10 instruction categories.

\paragraph{Dataset Composition.} We summarize the overall statistics of the dataset in Table~\ref{tab:data_stats_new}. The final benchmark contains 2,527 samples in total, spanning 5 disciplines and including ten functional constraint groups. Each sample consists of a verified question, optional visual input, and a reference answer. On average, each sample contains 211.03 tokens in the question. Among all samples, 27.50\% include multimodal inputs with associated images.

% 增加constraint distribution

\section{Experiment}
\label{sec:setup}
% We introduce the test closed-source and open-source models and multi-dimensional metrics in this section.
\subsection{Setup}

\textbf{Models.} To evaluate the effectiveness of our dataset and evaluation pipeline, we conduct experiments on a diverse set of state-of-the-art large language models (LLMs), including GPT-5.2~\cite{gpt5.2}, Grok-4-Fast~\cite{grok-4-fast}, Gemini-3.1-Pro-Preview~\cite{gemini}, and Claude-Sonnet-4.6~\cite{claude-sonnet-4-6}, which are widely recognized for their strong reasoning and instruction-following capabilities. Simultaneously, we also assess advanced open-source models, including InternVL3.5 series~\cite{internvl3.5}, and Qwen3.5 series~\cite{qwen3.5} to further analyze performance differences across model families. 

\noindent\textbf{Details.} For multimodal samples, images are preprocessed before generation to reduce transmission overhead during large-scale model inference while preserving essential semantic information. We set $N=3$, and $k=3$ to balance diversity and consistency in constraint augmentation.
%Overall, our evaluation covers 12 models across 2 model families, spanning both proprietary and open ecosystems.
%Detailed model configurations and inference settings are provided in Appendix~\ref{app:model_details}.  DeepSeek-V3.2-Speciale~\cite{deepseekv32}

\subsection{Metrics}

To evaluate model performance, we adopt several existing metrics that measure different aspects of instruction-following performance. Specifically, we employ the \textit{Constraint Satisfaction Rate} (CSR)~\cite{csr&isr} and the \textit{Instruction Satisfaction Rate} (ISR)~\cite{csr&isr}, as well as the \textit{Decomposed Requirements Following Ratio} (DRFR)~\cite{drfr}. 
Together, these metrics provide complementary perspectives on model performance, including CSR captures the averaged constraint-level satisfaction across samples, ISR evaluates instruction-level success, and DRFR measures decomposed-requirement-level compliance across the total number of constraints. Constraint verification is categorized into exact match, precision-based, and LLM-as-a-judge, with detailed protocols provided in the Appendix~\ref{app:evaluation}.
\section{Result \& Analysis}
\label{sec:result}
This section reports model performance across disciplines, constraint domains, and constraint groups. Subsequently, we analyze performance trends as the number of constraints increases and examine the relationship between answer correctness and instruction following. Due to space limitations, the main text focuses on
representative models and key findings. Additional analyses
of source-level variation,
modality-associated differences, correctness contrasts,
and representative cases are provided in the
Appendix~\ref{app:additional_results}.

\subsection{Performance on Different Science Disciplines }
% 我们首先根据数据来源所在的学科进行分类讨论，如表\ref{tab:main results}所示，我们在化学、地理、生物、材料和物理五个学科上，分别展示了模型指令跟随分数，以分析不同学科数据来源对模型的影响。
We first evaluate the instruction-following performance of MLLMs across different scientific disciplines. As summarized in Table \ref{tab:main results}, we report the CSR, ISR, and DRFR metrics for five domains to analyze how disciplinary data characteristics influence model behavior.

\textbf{Performance across disciplines.}
% 在生物和材料上面模型指令跟随能力最强，在化学和地理上指令跟随能力最差。从表\ref{tab:main results}中我们可以看出，xxx根据表格结果给出具体的数值说明。结合学科特点和学科指令分析为什么。生物和材料（模型对病例名称以及材料表征方法掌握比较好？xxx其他可能原因），化学和地理（对化学官能团、化学键识别、以及地理格式上训练不足？xxxx其他可能原因）
\textit{Models demonstrate that no single model performs well across all disciplines, showing significant variation in performance across domains.} The highest ISR scores are observed in the biology and materials scientific domain, while performance drops significantly in chemistry and geography. Specifically, GPT-5.2 achieves an ISR of 72.82\% in biology and 78.07\% in materials science. In contrast, chemistry presents the greatest challenge, with the highest ISR for GPT-5.2 falling to 46.33\%. This disparity is likely due to the nature of the tasks in each domain: biology and materials science tasks often rely on entity recognition and method restriction, which are well-represented in the training data. On the other hand, chemistry and geography require models to handle more complex structural constraints and spatial hierarchies, which are harder to capture through standard linguistic alignment.

\textbf{Closed-Source vs Open-Source.}
% 在所有学科上，闭源模型的性能都强于开源模型。从表\ref{tab:main results}中我们可以看出，结合数值进行分析，最强的闭源模型xxx分数是xxx，最强的开源模型xxx分数是xxx，闭源模型比开源模型能力强xxx%，说明xxxxx。
\textit{Closed source models maintain a consistent performance advantage over open source models across all disciplines.} The leading closed-source model, GPT-5.2, achieves an overall score of 65.67\%, exceeding the 57.39\% attained by the leading open-source model, Qwen3.5-397B-A17B. Although the gap narrows in disciplines such as physics, where Qwen3.5-397B-A17B achieves an ISR of 62.50\%, compared with 69.56\% for GPT-5.2, closed-source models demonstrate greater robustness in satisfying multifaceted scientific constraints. This superiority reflects the advantages conferred by high-quality data and advanced alignment strategies in proprietary models. 

\textbf{Impact of Model Scale.}
% 参数量增加并不能带来指令跟随能力的提升。从表中看出，qwen3.5模型从27B到122B，overall分数分别是xx，没有什么变化；internvl3.5从8B到38B，overall分数分别是xxx，没有提升，反而下降。说明限制模型指令跟随能力的不是参数量，而是xxxx
\textit{An increase in parameter size does not lead to a linear improvement in the instruction-following performance of models in scientific contexts.} For the InternVL3.5 series, the 8B model yields an overall score of 51.37\% which is nearly identical to the 51.64\% of the 38B version. A more significant trend appears in the Qwen3.5 family where the score of the 27B model is 51.37\% while the 122B version drops to 50.41\%. These results suggest that the primary bottleneck for scientific instruction-following is not raw computational power but rather the quality of domain-aware alignment. Scaling up parameters without targeted optimization of scientific instruction-tuning data may cause performance saturation or an alignment tax that hinders the ability to satisfy complex disciplinary rules.

\begin{figure}
    \centering
    \includegraphics[width=\linewidth]{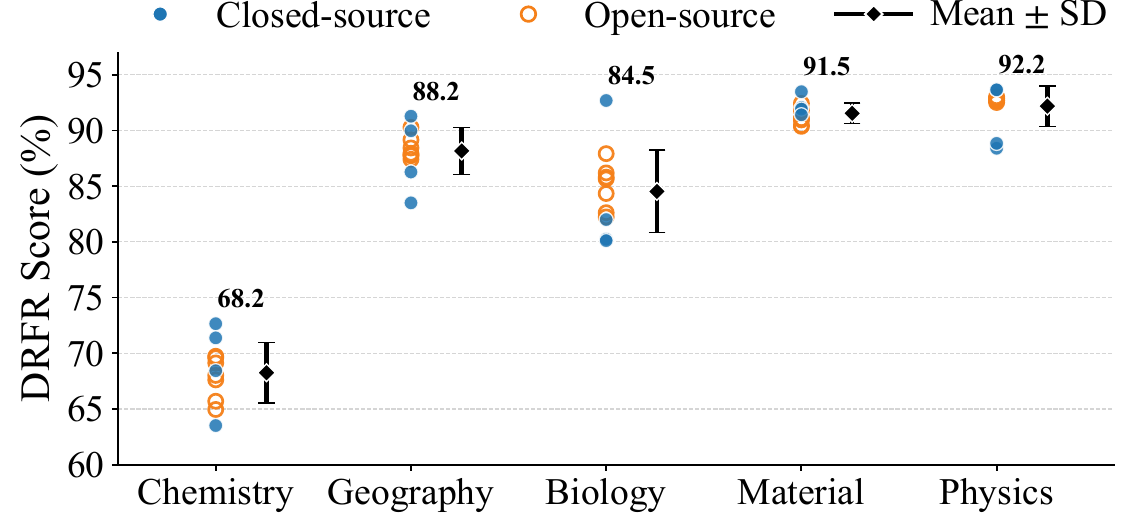}
    \caption{DRFR scores (\%) across different scientific constraint domains for the evaluated MLLMs.
}
    \label{fig:constraint domains}
\end{figure}

\begin{figure}
    \centering
    \includegraphics[width=\linewidth]{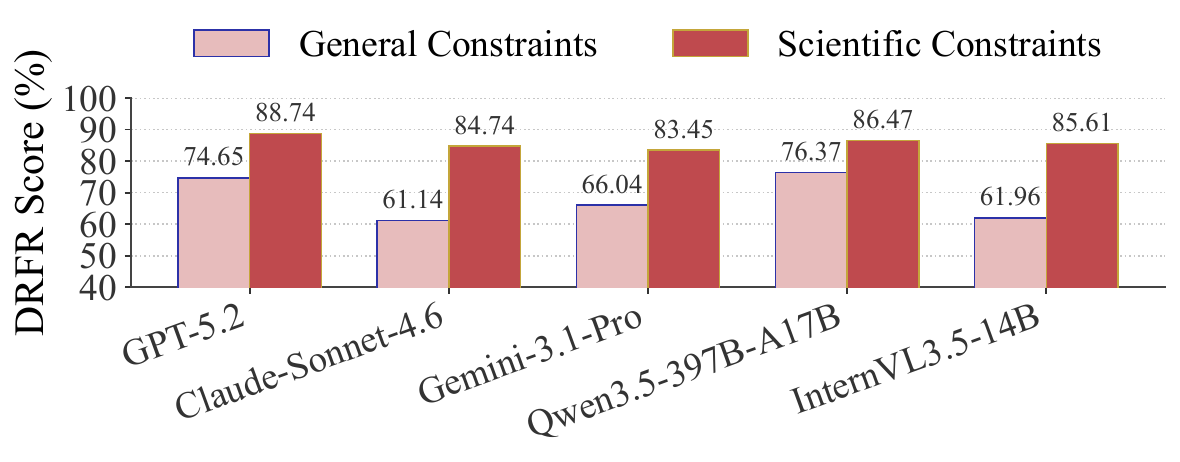}
    \caption{DRFR scores (\%) of representative MLLMs between general and scientific constraint domains.}
    \label{fig:general_vs_scientific}
\end{figure}

{

\begin{table*}[!h]
\centering
\small
\setlength{\tabcolsep}{2pt}

% \begin{adjustbox}{width=\linewidth}
% \renewcommand{\arraystretch}{1.2}
\begin{tabular*}{\textwidth}{@{\extracolsep{\fill}}l|cccccccccc@{}}
\hline

\textbf{Model}
& \textbf{Format}
& \textbf{Precision}
& \textbf{Selection} 
& \textbf{Structure}
& \textbf{Number}
& \textbf{Letter} 
& \textbf{Procedure}
& \textbf{Method}
& \textbf{Terminology}
& \textbf{Unit}
\\

\hline

% ===== 数据行示例=====
\multicolumn{11}{c}{\textit{Closed-Source MLLMs}}\\
\hdashline
\mbox{GPT-5.2} & 72.18 & 84.69 & 82.24 & \textbf{83.41} & \textbf{68.62} & 51.30 & \textbf{94.19} & 95.91 & 73.59 & 85.61 \\
\mbox{Grok-4-Fast} & \textbf{82.31} & 77.99 & 82.89 & 77.35 & 58.50 & \textbf{53.90} & 78.14 & 94.55 & 72.23 & 80.00\\
\mbox{Gemini-3.1-Pro-Preview} & 74.38 & 75.60 & 80.92 & 63.90 & 55.76 & 50.00 & 78.29 & 96.68 & 72.23 & 77.09\\
\mbox{Claude-Sonnet-4.6} & 50.67 & \textbf{85.37} & \textbf{83.44} & 76.91 & 58.89 & 35.33 & 76.74 & \textbf{98.28} & \textbf{74.60} & \textbf{88.86}\\
\hline
\multicolumn{11}{c}{\textit{Open-Source MLLMs}}\\
\hdashline
\mbox{Qwen3.5-27B} & 71.20 & 88.89 & 83.22 & 85.19 & 46.81 & 50.67 & 80.95 & 97.94 & \textbf{76.22} & 85.86 \\
\mbox{Qwen3.5-35B-A3B} & 72.53 & 87.23 & \textbf{84.21} & 85.55 & 43.23 & 52.00 & 80.70 & 98.16 & 74.64 & 84.57 \\
\mbox{Qwen3.5-122B-A10B} & 73.23 & 86.75 & 83.01 & 85.62 & 53.56 & \textbf{56.29} & 79.20 & \textbf{98.38} & 73.73 & 83.52\\
\mbox{Qwen3.5-397B-A17B} & \textbf{73.70} & \textbf{89.21} & 83.66 & \textbf{87.76} & \textbf{60.05} & 53.59 & 86.06 & 98.04 & 74.77 & 83.38\\
\mbox{InternVL3.5-8B} & 50.57 & 77.03 & 75.00 & 78.03 & 57.79 & 37.01 & 90.49 & 88.76 & 71.78 & 88.40 \\
\mbox{InternVL3.5-14B} & 53.63 & 81.82 & 76.32 & 80.27 & 57.56 & 36.36 & \textbf{90.94} & 90.03 & 69.53 & \textbf{88.69}  \\
\mbox{InternVL3.5-38B} & 48.76 & 81.34 & 77.63 & 76.46 & 56.88 & 38.31 & 89.49 & 90.55 & 71.11 & 87.52\\
% DeepSeek-V3.2-Speciale & 31.17 & 55.02 & \textbf{88.82} & 3.81 & 44.02 & 0.00 & 48.03 & \textbf{12.35} & \textbf{78.33} & \textbf{95.30}\\
\hline
\end{tabular*}
% \end{adjustbox}
\caption{DRFR scores (\%) of different constraint groups from evaluated MLLMs. The best score in each column is bolded separately for closed-source and open-source MLLMs. } 

\label{tab:constraint group}
\end{table*}
}

%\textbf{Declarative Memory} represents the average scores across explicit memory tasks, \textbf{Non-declarative} represents average scores across implicit memory tasks, and \textbf{Overall} represents the average performance of models between declarative memory and non-declarative memory. 

\subsection{Performance on Constraint Domains}
% 接下来我们从指令所属的domain来进行分析，一句话说明在这个分类下讨论的目的。
We explore performance across constraint domains to identify which requirements pose the greatest challenges to alignment. As established in the previous section, these domains include a general category and five scientific subcategories. 

\textbf{Performance across Scientific Constraint Domains.}
\textit{The distribution of constraint groups varies significantly across scientific disciplines, contributing to differing levels of task complexity.} As illustrated in Figure~\ref{fig:constraint domains}, chemistry is the most challenging domain, with an average DRFR score of only 68.2\% and  physics achieves the highest average performance with an average DRFR score of 92.\%. This variance is largely driven by the distinct constraint profiles inherent to each field. Chemistry tasks predominantly feature a high concentration of fine-grained constraints, demanding specific numerical values and strict structural formats. In contrast, domains like physics tend to rely more heavily on coarse-grained constraints. These usually involve methodological guidelines, broad analytical steps, or general reasoning procedures. Consequently, the dense distribution of highly specific, minute constraints in chemistry makes its tasks inherently more complex to execute than those in domains dominated by broader instructional constraints.

% 总体上genral比scientific更差，可能是因为scientific的约束可以帮助模型分析，完成问题本身的解答，所以模型更倾向于去遵循；而general指令，通常和模型正确回答无关，所以模型更倾向于忽略。
% \textit{Models achieve higher instruction-following performance in scientific domains than in general domains.} As shown in Figure~\ref{fig:general_vs_scientific}, GPT-5.2 achieves a DRFR of 88.74 in scientific, while Qwen3.5-397B-A17B achieves 86.47. In contrast, their performance in general domains is lower, with scores of 74.65 and 76.37, respectively. This gap may arise due to the nature of scientific constraints, which often involve reasoning methods and analysis steps that are integral to correctly answering the question. These structured constraints make it easier for models to follow and adhere to the task requirements. By contrast, general domain constraints may be less relevant to the solution and are therefore more likely to be overlooked. These findings suggest that stronger instruction alignment for scientific constraints could further improve MLLM performance.
\textbf{General vs Scientific Constraints.}
\textit{Models consistently perform better on scientific constraints than on general constraints.} As shown in Figure~\ref{fig:general_vs_scientific}, GPT-5.2 achieves a DRFR score of 88.74\% on scientific constraints, compared with 74.65\% on general constraints. A similar gap is observed for Qwen3.5-397B-A17B and other models. One possible explanation is that scientific constraints are often semantically coupled with the scientific task itself and models may therefore prefer to attend to these requirements. In contrast, general constraints primarily regulate the presentation and organization of the output and may be relatively independent of the underlying scientific reasoning process, making them more likely to be overlooked. These results highlight the need to evaluate both scientific and general constraints, as strong adherence to domain-specific requirements does not necessarily imply reliable compliance with the full instruction.

\subsection{Performance across Constraint Groups}

% 接下来我们从指令所属的group的角度来分析，如\label{sec:constraint_categories}所述，一共分为10个group，我们讨论模型在每个group上的表现。在Table \ref{tab:constraint group} 中展示了结果，汇报的DRFR指标。
We examine the performance of models across different functional groups to identify which instruction types are most manageable or challenging for current systems.  As detailed in section~\ref{sec:constraint_categories}, there are ten distinct groups and the results for the DRFR scores are summarized in Table~\ref{tab:constraint group}.

\textbf{Performance Consistency Across Groups.}
% 分析哪些group更难，以及更难的group上闭源和开源的差异是不是更大。甚至超过。
\textit{More challenging groups lead to more significant differences in model performance.} As observed, models struggle more in the number, letter, and format groups, with a large performance gap between GPT-5.2 and Qwen3.5-397B-A17B, the top-performing models in the closed-source and open-source categories, respectively. The difference in these groups can reach as high as 8\% to 9\%. However, in less challenging groups, such as precision, procedure, and method, the performance gap between models is much narrower, typically ranging from just 2\% to 4\%. This indicates that the group categorization is well-balanced and covers a range of difficulty levels, providing clear differentiation between model performance across various tasks.

\textbf{Analysis of the Challenge of Constraints.}
\textit{Models perform poorly on letter and number constraints, reflecting difficulties in both fine-grained symbolic processing and domain-specific knowledge application.} GPT-5.2 achieves DRFR scores of 51.30\% for letters, 68.62\% for numbers, and 73.59\% for terminology. Unlike generic sub-token tasks involving character counting or manipulation, our number constraints require models to interpret scientific structures and identify chemical bonds, functional groups, amino-acid residues, or sequence motifs rather than count numerical characters. Terminology constraints evaluate context-appropriate scientific term usage. These results highlight limited grounding of fine-grained symbols and quantities in scientific objects. Potential remedies include structure-aware representations, constrained decoding, and external scientific parsers or verification tools.

\subsection{Relationship Between Answer Correctness and Instruction-Following}
\begin{figure}
    \centering
    \includegraphics[width=\linewidth]{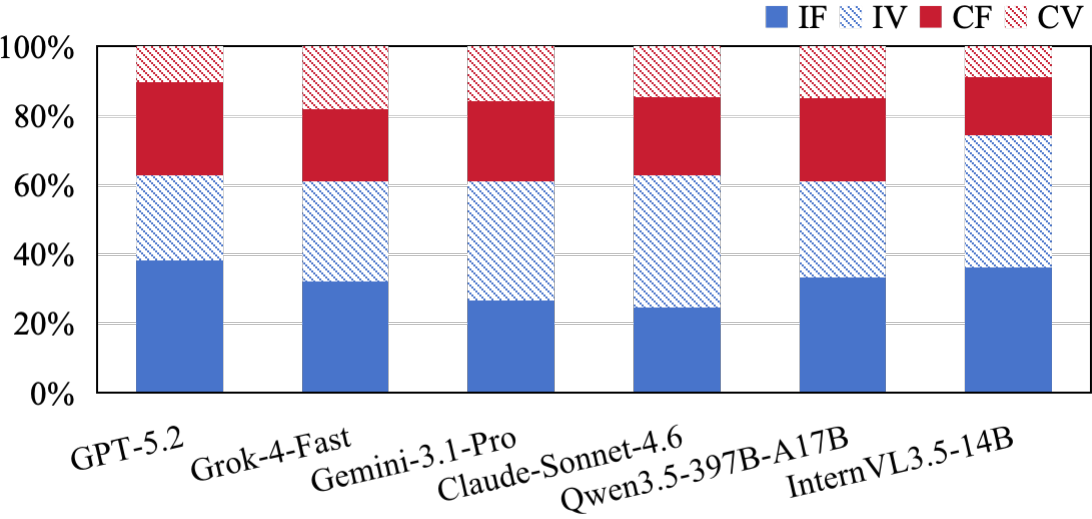}
    \caption{Proportions of Correct Followed (CF), Correct Violated (CV), Incorrect Followed (IF), and Incorrect Violated (IV) samples across representative models.}
    \label{fig:correct_vs_if}
\end{figure}

% 科学问题上做对+指令遵循，很难 2. 模型做对了，很大程度也会忽略指令，难以满足下游任务的需求 3. 很大一部分盲目的遵循指令 但是做错了，也是无效的。还有很大的改进空间。
We examine whether models can simultaneously produce scientifically correct answers and satisfy the specified constraints. Figure~\ref{fig:correct_vs_if} reports the proportion of samples in four categories: Correct and Followed (CF), Correct but Violated (CV), Incorrect but Followed (IF), and Incorrect and Violated (IV). Across all evaluated models, the CF rate remains below 30\%, indicating that jointly achieving scientific correctness and instruction adherence is still challenging. Moreover, approximately 20\% of samples fall into the CV category, showing that even correct answers frequently violate one or more constraints. The IF rate also exceeds 30\% for several models, suggesting that models may satisfy formatting or structural requirements without producing scientifically valid answers. These results reveal a clear gap between scientific correctness and constraint adherence, highlighting the need to improve both capabilities jointly rather than optimizing either in isolation. The verification procedures and weak association evidence are detailed in the Appendix~\ref{app:significance}.

% 一段话
% 这一部分我们讨论模型指令遵循能力和是否做对题目之间的关系。结果如figure~\cite{fig:correct_vs_if}所示，我们以4个闭源模型，和两个开源模型里表现最好的模型为代表，汇报了每个模型下，Correct–Followed (CF) 、Correct–Violated (CV) 、Incorrect–Followed (IF) 、Incorrect–Violated (IV)四种情况下的样本比例。我们可以从图中观察到，回答正确的比例（cf+cv），小于指令遵循的比例（cf+if）。给出数值佐证。说明模型回答正确比指令跟随更难。同时我们计算了每个模型的p-value，解释一下xxx， p值大于 0.05，表示答案正确性与指令遵循之间没有显著相关性。gemini-3.1-pro-preview是0.262，claude-sonnet-4-6 是 0.033，internvl3.5-14b是0.485，其他模型都是0。由此我们发现，显著相关的模型（GPT-5.2、Grok-4-fast、qwen3.5-397b-a17b、claude-sonnet-4-6）表现出强烈的 指令遵循与正确答案之间的关联，表明这些模型的答题准确率受到是否遵循指令的影响。不显著的模型（gemini-3.1-pro-preview、internvl3.5-14b）则没有明显的相关性，可能是因为这些模型在处理指令时的表现与答案准确度不太相关。

\section{Conclusion}
\label{sec:conclusion}
In this work, we propose SciMIF, a comprehensive benchmark for evaluating the instruction-following capabilities of MLLMs across five scientific disciplines: Chemistry, Geography, Biology, Material, and Physics. By systematically injecting 10 distinct groups of general and domain-specific constraints, SciMIF assesses a model's ability to seamlessly integrate rigorous disciplinary knowledge with complex formatting and methodological requirements. Our extensive evaluations show that the performance of MLLMs varies drastically by science domains and models are highly vulnerable to fine-grained constraints and suffer sharp performance drops as constraint counts increase. Our analysis also uncover a scaling paradox where merely increasing parameter size does not linearly improve scientific instruction following. 
% This benchmark provides a foundation for future research to improve scientific MLLMs for real-world applications.

Based on these findings, we suggest following directions for future research. First, training and evaluation should be reoriented toward practical application capabilities in constrained environments, rather than being primarily driven by scientific knowledge injection. Besides, to mitigate limitations related to symbolic and discrete constraints, MLLM-centric agents could consider integrating external numerical computation tools and symbolic engines to guide the inference process, enabling more precise and reliable control in real-world scientific workflows.

% In this work, we introduce SciMIF, a benchmark for evaluating multimodal scientific instruction-following across five disciplines and ten functional constraint groups. Our experiments reveal substantial performance variation across disciplines, persistent difficulties with fine-grained symbolic and quantitative constraints, and sharp degradation as the number of concurrent constraints increases. We further find that increasing model scale does not consistently improve scientific instruction-following. SciMIF nevertheless has several limitations: its current coverage is restricted to five disciplines and selected task types, while constraint injection over existing datasets may not fully capture naturally occurring instructions and long-horizon interactions in real scientific workflows. Future work should therefore extend SciMIF to broader disciplines, tasks, and interactive settings. Meanwhile, scientific MLLMs should be optimized for practical performance under constrained conditions rather than relying primarily on model scaling or knowledge injection, and may benefit from external numerical, symbolic, and scientific verification tools.

\newpage
\bibliography{main}
\clearpage

\appendix

%\section{Appendix Overview}

%The appendix is organized into seven sections:

%\begin{itemize}
%    \item Section~\ref{app:constraints} specifies the concrete constraint definitions over scientific and general domains. 
%    \item Section~\ref{app:construction_details} introduces the details of data construction including constraint recognition, injection, and human verification.
%    \item Section~\ref{app:data_sources} is involved in specific data source benchmarks and their task types. 
%    \item Section~\ref{app:evaluation} specifies both the complete definitions of three metrics and the evaluation protocols of correctness and instruction-following. 
%    \item Section~\ref{app:additional_results} showcases additional results and analysis to support the viewpoints in the paper, including complete performance over scientific and general constraints, analysis of the significance of correctness and instruction following, performance across source datasets, comparison on text-only and multimodel queries, analysis of the influence of injected instructions on correctness, error analysis on GPT-5.2 over different constraint groups, and representative case studies. 
%    \item Section~\ref{app:data_compliance} declares the compliance of the benchmark's source and that it will be publicly released.
%\end{itemize}

\section{Constraint Definitions}
\label{app:constraints}

\begin{table*}[htbp]
\centering
\begin{tabular}{lp{1.6cm}p{3.5cm}p{9.3cm}}
\toprule
\textbf{Domain} & \textbf{Group} & \textbf{Instruction Name} & \textbf{Description} \\
\midrule

\multirow[t]{4}{*}{General} & Format & json\_constraint & Format the output answer strictly as a JSON object.  \\
 & Selection & choose\_from & Select and output exactly one option from choices.\\
& Precision & decimal\_number & Ensure that all numerical quantities in the output retaining the required number of decimal places. \\
& Letter & all\_uppercase &  Output format must be in all uppercase.  \\
\cmidrule(lr){1-4}

\multirow[t]{2}{*}{Chemistry} & Terminology & molecular\_validity & Generate valid chemical nomenclature adhering to specified systems.  \\
& Procedure & reaction\_steps & Output an orderly, multi-step chemical reaction sequence. \\

\cmidrule(lr){1-4}

\multirow[t]{2}{*}{Geography}& Terminology & address\_hierarchy & Generate valid geographical addresses following spatial hierarchies. \\
& Unit & unit\_consistency &Output numerical results with specified geographical units.  \\

\cmidrule(lr){1-4}

\multirow[t]{2}{*}{Biology}& Method & method\_constraint & Use specified biological formulas or laws for calculation and reasoning.  \\
 & Numer & protein\_length & Output the required length of a specified biological sequence.  \\

\cmidrule(lr){1-4}

\multirow[t]{2}{*}{Material} & Terminology & characterization\_technique & Identify valid materials characterization techniques.  \\
 & Format & property\_type & Output different formats based on the material properties.  \\

\cmidrule(lr){1-4}

\multirow[t]{2}{*}{Physics} & Unit & unit\_consistency & Output numerical results using specified physical units. \\
& Procedure & analysis\_steps\_constraint & Solve problems step-by-step with physical principles. \\

\bottomrule
\end{tabular}
\caption{Representative Constraints Selected from Each Domain.}
\label{tab:constraints_overview}
\end{table*}

\subsection{Discipline-Specific Instantiations}
\label{app:discipline_instantiations}

\begin{figure}[t]
    \centering
    \includegraphics[width=\linewidth]{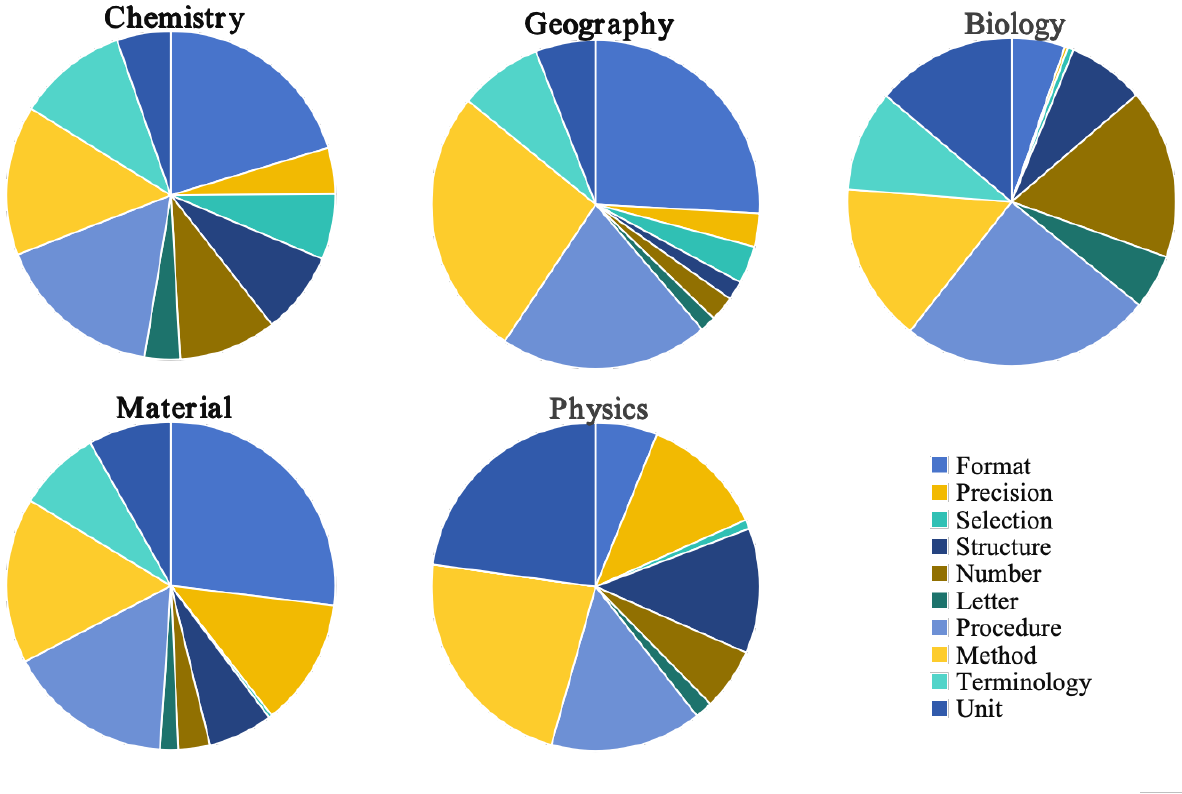}
    \caption{
    Distribution of the ten functional constraint groups across the
    five scientific disciplines in SciMIF. The proportions represent
    the frequency with which each group is instantiated within the
    corresponding discipline.
    }
    \label{fig:group_composition}
\end{figure}

Although the ten functional constraint groups are shared across SciMIF, their concrete meanings and implementations vary across scientific disciplines. This variation reflects differences in scientific representations, disciplinary conventions, reasoning processes, and output requirements. Figure~\ref{fig:group_composition} shows the distribution of functional constraint groups within each scientific discipline. Rather than applying generic instruction templates uniformly, SciMIF instantiates each group according to the knowledge and operational requirements of the corresponding domain.

\paragraph{General Constraints.}
General constraints provide cross-domain requirements that can be combined with scientific constraints from any discipline. They mainly regulate the representation and organization of model outputs, including output format, numerical precision, candidate selection, response structure, capitalization, and the required number of responses. For example, a model may be required to return a result as a JSON object, report all numerical quantities to a specified number of decimal places, select an answer from a predefined candidate set, or separate the reasoning process from the final conclusion. Although these constraints do not necessarily introduce additional scientific knowledge, they are important for ensuring that scientific outputs are unambiguous, machine-readable, and suitable for downstream automated evaluation or analysis.

\paragraph{Chemistry.}
Chemistry involves specialized molecular representations, stoichiometric relationships, and ordered reaction processes. Accordingly, terminology constraints require models to generate valid chemical nomenclature or molecular representations, such as SMILES, while preserving the identity and validity of the target molecule. Number constraints regulate scientifically meaningful quantities, including the  numbers of atoms, bonds, functional groups, or other molecular substructures. Procedure constraints require chemical transformations or synthesis routes to be expressed as ordered reaction steps. Method constraints may further specify the chemical principles, calculation rules, or analytical approaches that must be applied during reasoning. These instantiations evaluate whether models can satisfy explicit instructions while correctly interpreting molecular structure and chemical processes.

\paragraph{Geography.}
Geography emphasizes spatial organization, hierarchical relationships, scale dependence, and the interpretation of Earth-related observations. Terminology constraints require models to produce valid geographical expressions, such as locations organized according to administrative or spatial hierarchies. Format constraints are used when models must identify a geographical scene or return a result using a predefined category or structured representation. Unit constraints regulate geographical quantities such as distance, area, elevation, or radiation measurements. Procedure and method constraints specify the
analytical steps or spatial reasoning approaches that should be used to interpret maps, remote-sensing images, or other Earth science data. These constraints assess both spatial understanding and compliance with domain-specific geographical conventions.

\paragraph{Biology.}
Biology involves multi-level entities and processes ranging from molecular sequences to complex biological systems. Terminology constraints require models to use valid biological entity names or represent relationships among genes, proteins, molecules, and other biological entities. Number constraints specify quantities grounded in biological structures, such as protein lengths, sequence lengths, or the number of particular residues or motifs. Procedure constraints require models to organize biological analysis into a predefined sequence of operations, while method constraints require the use of specified biological formulas, analytical rules, or experimental approaches. These instantiations examine whether models can connect instruction with the structural and functional properties of biological systems.

\begin{table*}[htbp]
\centering
\begin{tabular}{p{1.5cm}p{5.3cm}p{7.8cm}p{1.8cm}}
\toprule
 \textbf{Group} & \textbf{Constraint Name} & \textbf{Description} & \textbf{Metric} \\
\midrule

 Unit & chemistry: unit\_consistency & Output numerical results using specified chemical stoichiometric units. & Script-Based \\
\cmidrule(lr){1-4}
 Terminology & chemistry: molecular\_validity & Generate valid chemical nomenclature adhering to specified systems. & Script-Based \\
\cmidrule(lr){1-4}
 Number & chemistry: atom\_count\_constraint & Ensure generated molecules contain a specified number of atoms. & Script-Based \\
  & chemistry: atom\_bond\_constraint & Ensure generated molecules contain a specified number of chemical bonds. & Script-Based \\
  & chemistry: atom\_group\_constraint & Ensure generated molecules contain a specified number of functional groups. & Script-Based \\
\cmidrule(lr){1-4}
 Method & chemistry: method\_constraint & Apply specified chemical formulas or laws for calculation and reasoning. & Script-Based \\
\cmidrule(lr){1-4}
 Procedure & chemistry: reaction\_steps & Output an orderly, multi-step chemical reaction sequence. & LLM-as-a-Judge \\

\bottomrule
\end{tabular}
\caption{Constraint Specifications in the Chemistry Domain.}

\label{tab:chemistry}
\end{table*}

\begin{table*}[htbp]
\centering
\begin{tabular}{p{1.5cm}p{5.3cm}p{7.8cm}p{1.8cm}}
\toprule
\textbf{Group} & \textbf{Constraint Name} & \textbf{Description} & \textbf{Metric} \\
\midrule

 Unit & geography: unit\_consistency & Output numerical results with specified geographical units. & Script-Based \\
\cmidrule(lr){1-4}
Terminology & geography: address\_hierarchy & Generate valid geographical addresses following spatial hierarchies. & Script-Based \\
\cmidrule(lr){1-4}
 Format & geography: scene\_option\_constraint & Select the answer strictly from a provided list of geographical scenes. & Script-Based \\
\cmidrule(lr){1-4}
 Method & geography: method\_constraint & Apply specified geographical formulas or laws for calculation and reasoning. & Script-Based \\
\cmidrule(lr){1-4}
 Procedure & geography: analysis\_steps\_constraint & Conduct reasoning according to specified geographical analysis steps. & LLM-as-a-Judge \\

\bottomrule
\end{tabular}
\caption{Constraint Specifications in the Geography Domain.}
\label{tab:geography}
\end{table*}

\begin{table*}[htbp]
\centering
\begin{tabular}{p{1.5cm}p{5.3cm}p{7.8cm}p{1.8cm}}
\toprule
 \textbf{Group} & \textbf{Constraint Name} & \textbf{Description} & \textbf{Metric} \\
\midrule

 Unit & biology: unit\_consistency & Output numerical results using specified biological units. & Script-Based \\
\cmidrule(lr){1-4}
 Terminology & \raggedright biology: entity\_\allowbreak relationship\_\allowbreak format\_\allowbreak validity & Extract and format specified biological entity relationships accurately. & Script-Based \\
\cmidrule(lr){1-4}
 Number & biology: protein\_length & Output the required length of a specified biological sequence. & Script-Based \\
\cmidrule(lr){1-4}
 Method & biology: method\_constraint & Use specified biological formulas or laws for calculation and reasoning. & Script-Based \\
\cmidrule(lr){1-4}
 Procedure & biology: analysis\_steps\_constraint & Analyze biological molecules according to a specified sequential bio-analysis flow. & LLM-as-a-Judge \\

\bottomrule
\end{tabular}
\caption{Constraint Specifications in the Biology Domain.}
\label{tab:biology}
\end{table*}

\begin{table*}[htbp]
\centering
\begin{tabular}{p{1.5cm}p{5.3cm}p{7.8cm}p{1.8cm}}
\toprule
\textbf{Group} & \textbf{Constraint Name} & \textbf{Description} & \textbf{Metric} \\
\midrule

 Unit & material: unit\_consistency & Output numerical results using specified materials science units. & Script-Based \\
\cmidrule(lr){1-4}
 Terminology & material: characterization\_technique & Identify valid materials characterization techniques. & Script-Based \\
\cmidrule(lr){1-4}
 Format & material: property\_type & Output different formats based on the material properties. & Script-Based \\
\cmidrule(lr){1-4}
 Method & material: method\_constraint & Apply specified materials science formulas or laws for reasoning. & Script-Based \\
\cmidrule(lr){1-4}
 Procedure & material: analysis\_steps\_constraint & Conduct reasoning according to specified materials analysis steps. & LLM-as-a-Judge \\

\bottomrule
\end{tabular}
\caption{Constraint Specifications in the Material Domain.}

\label{tab:material}
\end{table*}

\begin{table*}[htbp]
\centering
\begin{tabular}{p{1.5cm}p{5.3cm}p{7.8cm}p{1.8cm}}
\toprule
 \textbf{Group} & \textbf{Constraint Name} & \textbf{Description} & \textbf{Metric} \\
\midrule

 Unit & physics: unit\_consistency & Output numerical results using specified physical units. & Script-Based \\
\cmidrule(lr){1-4}
 Method & physics: method\_constraint & Apply specified physical formulas or laws for calculation and reasoning. & Script-Based \\
\cmidrule(lr){1-4}
 Procedure & physics: analysis\_steps & Solve problems step-by-step with physical principles. & LLM-as-a-Judge \\

\bottomrule
\end{tabular}
\caption{Constraint Specifications in the Physics Domain.}

\label{tab:physics}
\end{table*}

\begin{table*}[htbp]
\centering
\begin{tabular}{p{1.5cm}p{5.3cm}p{7.8cm}p{1.8cm}}
\toprule
 \textbf{Group} & \textbf{Constraint Name} & \textbf{Description} & \textbf{Metric} \\
\midrule

  Precision & general: decimal\_number & Ensure that all numerical quantities in the output retaining the required number of decimal places. & Script-Based \\
  & general: scientific\_annotation & Express all numerical quantities strictly in scientific notation. & Script-Based \\
  \cmidrule(lr){1-4}
  Letter & general: all\_uppercase & Output format must be in all uppercase. & Script-Based \\
  & general: all\_lowercase & Format the output answer strictly in all lowercase letters. & Script-Based \\
  \cmidrule(lr){1-4}
  Format & general: wrap\_up & Enclose the final output answer within a specified box format. & Script-Based \\
   & general: json\_constraint & Format the output answer strictly as a JSON object. & Script-Based \\
   & general: list\_constraint & Format the output answer strictly as a Python list. & Script-Based \\
   & general: tuple\_constraint & Format the output answer strictly as a Python tuple. & Script-Based \\
   & general: dictionary\_constraint & Format the output answer strictly as a Python dictionary. & Script-Based \\
   & general: markdown\_constraint & Format the output answer strictly using Markdown syntax. & Script-Based \\
   & general: html\_constraint & Format the output answer strictly using HTML tags. & Script-Based \\
   & general: xml\_constraint & Format the output answer strictly using XML tags. & Script-Based \\
   & general: csv\_constraint & Format the output answer as Comma-Separated Values (CSV). & Script-Based \\
  \cmidrule(lr){1-4}
  Selection & general: choose\_from & Select and output exactly one option from choices. & Script-Based \\
   & general: judge & Output strictly ``Yes'' or ``No'' without additional explanations. & Script-Based \\
  \cmidrule(lr){1-4}
  Structure & general: response\_structure & Follow a structured framework placing final answer at a specified position. & Script-Based \\

  \cmidrule(lr){1-4}
  Number & general: number\_response & Generate a specific number ($N$) of distinct categorical responses. & LLM-as-a-Judge \\

\bottomrule

\end{tabular}
\caption{Constraint Specifications in the General Domain.}
\label{tab:general}
\end{table*}

\paragraph{Material.}
Material relies heavily on multimodal characterization, property representation, experimental procedures, and structure-property relationships. Terminology constraints require models to identify valid characterization techniques based on textual and visual evidence. Format constraints specify how material properties should be represented, for example as discrete categories, continuous values, or structured labels. Procedure constraints regulate the ordered analysis of experimental observations or characterization results, while method constraints require reasoning to follow designated material principles or analytical approaches. Number and unit constraints may additionally regulate material quantities such as composition, mass change, dimensional measurements, or physical properties. These constraints evaluate whether models can integrate multimodal evidence with appropriate materials knowledge and report the result in the requested form.

\paragraph{Physics.}
Physics emphasizes dimensional consistency, causal deduction, and principle-based reasoning. Unit constraints require numerical quantities to be expressed using specified physical units while preserving
dimensional consistency. Method constraints require the explicit application of designated physical laws, equations, or theorems during problem solving. Procedure constraints specify the sequence of modeling, derivation, calculation, and verification steps that should be followed. %Precision and format constraints may further regulate how intermediate or final physical quantities are reported. 
These instantiations assess whether models can follow explicit analytical requirements while maintaining logically valid and physically consistent derivations.

% Overall, these discipline-specific instantiations preserve a common
% functional interpretation for each constraint group while grounding
% individual constraints in the representations, entities, conventions,
% and reasoning practices of each scientific field. This design supports
% both within-discipline evaluation and cross-disciplinary comparison
% along shared capability dimensions. Representative constraints are
% summarized in Table~\ref{tab:constraints_overview}, and the complete
% constraint definitions and evaluation methods are provided in
% Tables~\ref{tab:general}, \ref{tab:chemistry}, \ref{tab:geography},
% \ref{tab:biology}, \ref{tab:material}, and \ref{tab:physics}.

% \subsection{Complete Constraint Specifications}

% % This section first provides an overview of representative constraints from each domain in Table~\ref{tab:constraints_overview}. We then present the name, description, and evaluation method of each constraint in Tables~\ref{tab:chemistry}, \ref{tab:geography}, \ref{tab:biology}, \ref{tab:material}, \ref{tab:physics}, and \ref{tab:general}.

% \input{tables/constraints_overview}

\subsection{Complete Constraint Specifications}
\label{app:complete_constraints}

Table~\ref{tab:constraints_overview} presents representative constraints selected from the general domain and the five scientific disciplines. The examples illustrate how the functional constraint groups are instantiated as concrete and independently evaluable requirements in different scientific contexts. For each constraint, we report its functional group, canonical name, and operational description.

The complete constraint inventory contains 42 individual constraints organized into ten functional groups. Each constraint is assigned a canonical identifier in the form \texttt{domain:constraint\_name}, where \texttt{domain} specifies whether the constraint belongs to the general domain or one of the five scientific disciplines. This naming convention distinguishes individual constraints that belong to the same functional group but require different disciplinary knowledge or evaluation procedures. For example, the terminology group includes molecular validity in Chemistry, address hierarchy in Geography, biological entity relationships in Biology, and characterization techniques in Material.

Each constraint specification contains four components: (1) the functional group to which the constraint belongs; (2) a canonical constraint name; (3) an operational description defining the requirement imposed on
the model output; and (4) an evaluation method used to determine whether the requirement is satisfied. 
For constraints with deterministic output conditions, a rule-based evaluation approach is adopted, including output format, numerical precision, capitalization, units, candidate selection, and the number of specified scientific entities. Constraints requiring the assessment of open-ended reasoning processes, such as the application of a designated scientific method or adherence to a specified analytical procedure, are evaluated using an LLM-as-a-Judge protocol.

The complete chemistry constraint specifications are provided in Table~\ref{tab:chemistry}. They cover requirements involving chemical units, molecular representations, atom and bond quantities, functional groups, designated calculation methods, and ordered reaction procedures.

The complete geography constraint specifications are provided in Table~\ref{tab:geography}. They include geographical units, spatial and administrative hierarchies, geographical scene representations, designated spatial-analysis methods, and ordered analysis procedures.

The complete biology constraint specifications are provided in Table~\ref{tab:biology}. They cover biological units, entity relationships, sequence-related quantities, designated biological methods, and sequential bio-analysis procedures.

The complete material constraint specifications are provided in Table~\ref{tab:material}. They include materials science units, characterization techniques, material-property representations, designated analytical methods, and ordered materials analysis procedures.

The complete physics constraint specifications are provided in Table~\ref{tab:physics}. They cover physical units, designated physical laws and formulas, and step-by-step derivation or analysis procedures.

Finally, the complete general constraint specifications are provided in Table~\ref{tab:general}. These constraints can be combined with discipline-specific requirements and primarily regulate output format,
numerical precision, candidate selection, response structure, capitalization, and the required number of output items.

Together, these specifications provide an explicit mapping from the constraint taxonomy to discipline-grounded and independently verifiable requirements. They enable model performance to be evaluated
both at the level of individual scientific requirements and across common instruction-following capability groups.

\FloatBarrier

%%%%%%%%%%%%%%%%%%%%%%%%%%%%%%%%%%%%%%%%%%%%%%%%%%%%%%%%%%%%%%%%%%%

\section{Detailed Data Construction Procedure}
\label{app:construction_details}

This section provides the formal definitions and implementation details of the SciMIF construction pipeline described in Section Data Construction. An overview of the complete procedure is provided in Algorithm~\ref{alg:data-construction}.

\subsection{Seed Representation and Applicable Constraints}

Each seed sample from a source dataset is represented as
\begin{equation}
    x = (q, I, a, t),
\end{equation}
where $q$ is the original textual query, $I$ is an optional visual input, $a$ is the reference answer, and $t$ denotes the scientific task type.

Let $\mathcal{C}$ denotes the complete constraint inventory of SciMIF. For each sample $x$, we construct an applicable subset
\begin{equation}
    \mathcal{C}_x
    =
    \mathcal{C}^{\mathrm{sci}}_x
    \cup
    \mathcal{C}^{\mathrm{gen}}_x,
\end{equation}
where $\mathcal{C}^{\mathrm{sci}}_x$ contains scientific constraints compatible with the discipline and task type of $x$, and $\mathcal{C}^{\mathrm{gen}}_x$ contains general constraints applicable to the sample $x$.

\subsection{Existing Constraint Filtering}

Annotators manually identify constraints, including scientific and general, that are already explicitly stated or implicitly required by the original query. The recognized constraint set is defined as
\begin{equation}
    C_o
    =
    f_r(q, C_x),
\end{equation}
where $f_r$ denotes the manual recognition process, and the set $\mathcal{C}_o \subseteq \mathcal{C}_x$.

Recognizing existing constraints before augmentation prevents the construction process from repeatedly injecting an equivalent requirement or introducing a new constraint that conflicts with the original task.

\subsection{Candidate Constraint Selection}

The initial candidate pool is constructed from $C_x \setminus C_o$, in which the applicable constraints that are not already present in the query.

A selection operation is then applied:
\begin{equation}
    (C_s, C_g)
    =
    f_s
    \left(
        x,
        C_x \setminus C_o
    \right),
\end{equation}
where $C_s$ represents the scientific constraints that domain experts determine are compatible with the discipline, task type $t$, visual input $I$, and reference answer $a$. $C_g$ represents general constraints that annotators exclude constraints that duplicate existing requirements or conflict with either the original query or the selected scientific constraints. 

\subsection{Scientific Constraint Injection}

The selected scientific constraints are first adapted to the original query:
\begin{equation}
    q_d
    =
    f_d
    \left(
        q,
       C_s
    \right),
\end{equation}
where $q_d$ denotes the query after scientific constraint injection. %and
%\begin{equation}
%    \mathcal{C}_2 =\mathcal{C}_1 \cup\mathcal{C}^{\mathrm{sci}}_{\mathrm{inj}}.
%\end{equation}
%Here, $\mathcal{C}^{\mathrm{sci}}_{\mathrm{inj}}$ denotes the scientific constraints successfully incorporated into $q_d$.

During adaptation, the wording of each scientific constraint may be adjusted to match the terminology and conventions of the corresponding discipline. However, the required scientific operation and evaluation criterion remain unchanged.

\subsection{General Constraint Injection}

%Let \(\{G_1,\ldots,G_L\}\), where \(L\leq N\), denote the sampled general-constraint categories. Each category \(G_i\) contains a randomly ordered sequence of candidate constraints \(\{c_{i,1},\ldots,c_{i,M_i}\}\). At most one constraint is accepted from each category.

\begin{algorithm}[t]
\caption{SciMIF Data Construction}
\label{alg:data-construction}
\begin{algorithmic}[1]
\Require Seed sample $x=(q,I,a,t)$, constraint inventory $\mathcal{C}$, $N$, $k$
\Ensure Augmented sample $\hat{x}$

\State Construct applicable constraints
$\mathcal{C}_x=\mathcal{C}^{\mathrm{sci}}_x
\cup\mathcal{C}^{\mathrm{gen}}_x$
\State Identify existing constraints
$\mathcal{C}_o=f_r(q,\mathcal{C}_x)$
\State Select compatible constraints
$(C_s,C_g)=f_s(x,\mathcal{C}_x\setminus\mathcal{C}_o)$
\State Inject scientific constraints
$q_d=f_d(q,C_s)$
\State Sample at most $N$ categories from $C_g$
\State $q^{(0)}\gets q_d$

\For{$i=1,\ldots,N$}
    \State $q^{(i)}\gets q^{(i-1)}$
    \ForAll{candidate constraints $c_n$ in category $i$}
        \State Generate
        $q_g=f_g(q^{(i-1)},c_n)$ up to $k$ times
        \If{$\mathbb{I}_{\mathrm{included}}(q_g,c_n)
        \land\mathbb{I}_{\mathrm{valid}}(q_g,a)$}
            \State $q^{(i)}\gets q_g$
            \State \textbf{break}
        \EndIf
    \EndFor
\EndFor

\State $\hat q\gets q^{(N)}$
\State $\hat{x}\gets(\hat q,I,a,t)$
\State \Return $\hat{x}$
\end{algorithmic}
\end{algorithm}

%Thus, the augmentation is performed sequentially across categories: the accepted output from category \(G_i\) becomes the input to category \(G_{i+1}\). If no candidate in a category passes verification, the query remains unchanged for that category.

We sample at most $N$ mutually compatible general constraint categories from $C_g$. Let the selected constraints be
\begin{equation}
    \{C_g^1, C_g^2,  \ldots, C_g^N\},
    %\qquad
    %c_n \in \mathcal{C}^{\mathrm{gen}}_s.
\end{equation}
Each category contains multiple constraints, one of which is $c_n$. General constraints are manually categorized according to their compatibility, and at most one constraint is selected from the same category for a single augmented instruction.

$c_n$ is injected sequentially, then the candidate augmented query is generated as
\begin{equation}
    q_g= f_g\left(
        q_d,
        c_n
    \right).
\end{equation}

The candidate query is accepted only when it passes both automatic checks:
\begin{equation}
    \mathbb{I}_{{included}}
    \left(
        q_g, c_n
    \right)
    \land
    \mathbb{I}_{{unchanged}}
    \left(
       q_g, a
    \right)
    = 1.
\end{equation}
$I_{{included}}$ verifies that the target constraint is explicitly and unambiguously expressed in the augmented query, while $I_{unchanged}$ verifies that the original reference answer remains semantically sufficient for the augmented query. Literal equality is not required, which means equivalent numerical representations, capitalization changes, and output formatting are permitted, provided that the factual content of the answer remains correct and sufficient. We utilize DeepSeek-Chat as the verification model.

%If both checks are satisfied, the query and constraint set are updated as
%\begin{equation}
%    q^{(n)} = \widetilde{q}^{(n)},
%    \qquad
%    \mathcal{C}^{(n)}
%    =
%    \mathcal{C}^{(n-1)}
%    \cup
%    \{c_n\}.
%\end{equation}

If either check fails, the generation is retried up to $k$ times. When all $k$ attempts for $c_n$ fail, another compatible constraint from the same category is selected and verified using the same procedure. If no constraint in that category passes the verification, the category is skipped rather than forcing an invalid augmentation.

After sequentially processing the selected general constraints, the final augmented query and complete constraint set are defined as $\hat{q}$. The resulting augmented sample is
\begin{equation}
    \hat{x}
    =
    (\hat{q}, I, a, t).
\end{equation}

{

\begin{table*}[!h]
\centering

\begin{tabular}{p{0.18\textwidth} |
  p{0.18\textwidth}
  p{0.04\textwidth} |
  p{0.23\textwidth} |
  p{0.19\textwidth}
  p{0.04\textwidth}}
\hline

\textbf{Source}
& \textbf{Task}
& \textbf{Num.}
& \textbf{Source}
& \textbf{Task}
& \textbf{Num.} \\
\hline

\makecell[l]{ChemEval\\[-2pt]{\footnotesize\cite{chemeval}}}
& \makecell[l]{chemistry\_numerical\\\_task} & 70
& \makecell[l]{IMAGEO-Bench\\[-2pt]{\footnotesize\cite{IMAGEO-Bench}}}
& earth\_scene\_option & 100 \\

\makecell[l]{ChemEval\\[-2pt]{\footnotesize\cite{chemeval}}}
& chemistry\_entity\_option & 70
& \makecell[l]{LAB-Bench\\[-2pt]{\footnotesize\cite{lab-bench}}}
& biology\_sequence\_qa & 143 \\

\makecell[l]{ChemEval\\[-2pt]{\footnotesize\cite{chemeval}}}
& chemistry\_reaction\_steps & 20
& \makecell[l]{Mol-Instructions\\[-2pt]{\footnotesize\cite{Mol-Instructions}}}
& \makecell[l]{biology\_entity\_\\relationship} & 100 \\

\makecell[l]{ChemEval\\[-2pt]{\footnotesize\cite{chemeval}}}
& \makecell[l]{chemistry\_molecular\\\_format} & 70
& \makecell[l]{Mol-Instructions\\[-2pt]{\footnotesize\cite{Mol-Instructions}}}
& biology\_analysis\_steps & 250 \\

\makecell[l]{ChemEval\\[-2pt]{\footnotesize\cite{chemeval}}}
& chemistry\_open\_ended & 198
& \makecell[l]{MaScQA\\[-2pt]{\footnotesize\cite{mascqa}}}
& \makecell[l]{material\_numerical\\\_problem} & 100 \\

\makecell[l]{S2-TOMG-Bench-mini\\[-2pt]{\footnotesize\cite{TOMG-Bench}}}
& \makecell[l]{chemistry\_molcustom\_\\bondnum} & 30
& \makecell[l]{MatCha\\[-2pt]{\footnotesize\cite{matcha}}}
& \makecell[l]{material\_characterization\\\_technique} & 100 \\

\makecell[l]{S2-TOMG-Bench-mini\\[-2pt]{\footnotesize\cite{TOMG-Bench}}}
& \makecell[l]{chemistry\_molcustom\_\\functionalgroup}
 & 30
& \makecell[l]{LLM4Mat-Bench\\[-2pt]{\footnotesize\cite{llm4mat}}}
& \makecell[l]{material\_property\\\_prediction} & 97 \\

\makecell[l]{S2-TOMG-Bench-mini\\[-2pt]{\footnotesize\cite{TOMG-Bench}}}
& \makecell[l]{chemistry\_molcustom\_\\atomnum} & 30
& \makecell[l]{MatSciBench\\[-2pt]{\footnotesize\cite{matscibench}}}
& \makecell[l]{material\_reasoning\\\_process} & 200 \\

\makecell[l]{EarthSE\\[-2pt]{\footnotesize\cite{earthse}}}
& earth\_calculation\_task & 73
& \makecell[l]{UGPhysics\\[-2pt]{\footnotesize\cite{ugphysics}}}
& \makecell[l]{physics\_formatted\_\\numerical\_task} & 150 \\

\makecell[l]{EarthSE\\[-2pt]{\footnotesize\cite{earthse}}}
& earth\_reasoning\_steps & 250
& \makecell[l]{PhysReason\\[-2pt]{\footnotesize\cite{physreason}}}
& \makecell[l]{physics\_formatted\_\\reasoning\_process} & 196 \\

\makecell[l]{IMAGEO-Bench\\[-2pt]{\footnotesize\cite{IMAGEO-Bench}}}
& earth\_address\_format & 100
& \makecell[l]{PhysUniBench\\[-2pt]{\footnotesize\cite{phyunibench}}}
& \makecell[l]{physics\_formatted\_\\open\_ended} & 150 \\

\hline

\end{tabular}

\caption{Overview of task types, source datasets, and sample counts. } 
\label{tab:task type}
\end{table*}
}

\subsection{Human Verification}
\label{app:human_verification}

Automatic validation ensures that the target constraints are explicitly included in the augmented query and that the original reference answer remains valid. However, automatic rules may not fully capture linguistic
naturalness, implicit contradictions, or ambiguous scientific expressions. We therefore conduct human verification on all augmented samples before including them in the final benchmark. Two annotators independently review each augmented query $\hat{q}$ together with its associated constraint set. The verification focuses on the following two criteria.

\paragraph{Logical Coherence and Fluency.}
The augmented query must remain semantically coherent and grammatically fluent after constraint injection. Each injected constraint should be naturally integrated into the original scientific problem and should not introduce contradictory requirements, ambiguous references, unnecessary repetition, or incompatibility with
the original task setting.

\paragraph{Constraint Fidelity.}
Every constraint must be explicitly and unambiguously expressed in $\hat{q}$. The augmented query must preserve the intended operational meaning and evaluation condition of each constraint, rather than merely mentioning related terminology. The annotators also verify that the injected requirements do not alter the
scientific question, change the expected answer, or introduce additional conditions that cannot be satisfied by the original reference answer.

\paragraph{Annotation Consistency.}
Before full-scale verification, the two annotators independently examined the same 20 randomly sampled instances to calibrate the verification criteria. They reached the same decision on all sampled instances, yielding an observed agreement of 100\%. This pilot verification was used to confirm that the two criteria could be applied consistently before the remaining samples were reviewed.

\paragraph{Verification Results.}
Samples that fail either criterion undergo manual revision. Annotators rewrite only the problematic portions of the augmented query while preserving the original scientific task, optional visual input, and
reference answer. When a sample cannot be repaired without changing its scientific meaning or making one or more constraints invalid, it is discarded. All retained samples undergo human verification, and 884
samples are manually revised before inclusion in the final benchmark $\mathcal{D}'$.

%%%%%%%%%%%%%%%%%%%%%%%%%%%%%%%%%%%%%%%%%%%%%%%%%%%%%%%%%%%%%%%%%%%

\section{Data Sources and Task Coverage}
\label{app:data_sources}

SciMIF is constructed from 13 existing scientific datasets spanning five disciplines. The chemistry sources include S2-TOMG-Bench-mini~\cite{TOMG-Bench} and ChemEval~\cite{chemeval}; the geography sources include EarthSE~\cite{earthse} and IMAGEO-Bench~\cite{IMAGEO-Bench}; the biology sources include Mol-Instructions~\cite{Mol-Instructions} and Lab-Bench~\cite{lab-bench}; the material sources include MaScQA~\cite{mascqa}, MatCha~\cite{matcha}, LLM4Mat-Bench~\cite{llm4mat}, and MatSciBench~\cite{matscibench}; and the physics sources include UGPhysics~\cite{ugphysics}, PhysReason~\cite{physreason}, and PhysUniBench~\cite{phyunibench}.

Together, these datasets contribute 22 task types with diverse problem formulations, disciplinary conventions, and input modalities. Specifically, SciMIF contains eight task types in Chemistry, four in Geography, three in Biology, four in Material, and three in Physics. The complete task list, corresponding data sources, and sample statistics are presented in Table~\ref{tab:task type}. This broad coverage increases the structural diversity of SciMIF and reduces its dependence on any single discipline or task design.

\section{Evaluation Details}
\label{app:evaluation}

\subsection{Instruction-Following Metrics}
In this section, we introduce specific calculation formula of each metric as follows.

 \textbf{CSR} measures the average proportion of satisfied constraints across all instructions. Formally, it is defined as
 \[
 \mathrm{CSR} = \frac{1}{m}\sum_{i=1}^{m}\left(\frac{1}{n_i}\sum_{j=1}^{n_i}s_{i,j}\right),
 \]
 where $m$ denotes the total number of instructions, $n_i$ represents the number of constraints associated with the $i$-th instruction, and $s_{i,j} \in \{0,1\}$ indicates whether the $j$-th constraint in the $i$-th instruction is satisfied.

 \textbf{ISR} evaluates the proportion of instructions for which all associated constraints are completely satisfied. It is computed as
 \[
 \mathrm{ISR} = \frac{1}{m}\sum_{i=1}^{m}s_i,
 \]
 where $s_i \in \{0,1\}$ indicates whether all constraints in the $i$-th instruction are satisfied.

 \textbf{DRFR} measures the overall satisfaction of decomposed requirements across all instructions. Instead of evaluating instructions as a whole, this metric assesses requirement-level compliance through a set of scoring questions. It is defined as
 \[
 \mathrm{DRFR} =
 \frac{\sum_i \sum_{j=1}^{m_i} r'_{i,j}}
{\sum_i m_i},
 \]
where $m_i$ denotes the number of scoring questions associated with the $i$-th instruction, and $r'_{i,j}$ represents the result of the $j$-th scoring question for the $i$-th instruction.

{

\begin{table*}[!t]
\centering
\small

% \begin{adjustbox}{width=\linewidth}
% \renewcommand{\arraystretch}{1.2}
\begin{tabular}{l|ccccc|cc}
\hline

\textbf{Model}
& \textbf{Chemistry}
& \textbf{Geography}
& \textbf{Biology} 
& \textbf{Material}
& \textbf{Physics}
& \textbf{General} 
& \textbf{Scientific}

\\

\hline

% ===== 数据行示例=====
\multicolumn{8}{c}{\textit{Closed-Source MLLMs}}\\
\hdashline
%GPT-5.2 & \textbf{46.92} & 56.73 & \textbf{73.91} & 63.94 & \textbf{57.23} & \textbf{75.52} & \textbf{59.84} \\
GPT-5.2 & \textbf{72.65} & \textbf{91.28} & \textbf{92.69} & 92.06 & \textbf{93.66} & \textbf{74.65} & \textbf{88.74} \\
%Grok-4-Fast & 43.25 & 53.44 & 62.54 & 65.37 & 54.35 & 73.69 & 55.76\\
Grok-4-Fast & 68.43 & 83.50 & 82.00 & 91.92 & 88.40 & 74.46 & 82.93\\
%Gemini-3.1-Pro-Preview & 39.61 & \textbf{57.90} & 62.48 & \textbf{66.66} & 52.79 & 66.59 & 55.78\\
Gemini-3.1-Pro-Preview & 63.51 & 89.97 & 80.24 & \textbf{93.48} & 88.84 & 66.04 & 83.45\\
Claude-Sonnet-4.6 & 71.38 & 86.27 & 80.11 & 91.41 & 93.65 & 61.14 & 84.74\\
\hline
\multicolumn{8}{c}{\textit{Open-Source MLLMs}}\\
\hdashline
%Qwen3.5-27B & 41.94 & 55.91 & 60.64 & 63.34 & 55.88 & 73.86 & 55.33  \\
Qwen3.5-27B & 67.61 & 87.73 & 85.81 & 91.04 & 92.91 & 74.34 & 85.26  \\
%Qwen3.5-35B-A3B & 39.87 & 55.81 & 61.25 & 62.73 & 55.45 & 74.20 & 54.61 \\
Qwen3.5-35B-A3B & 65.69 & 87.95 & 86.19 & 91.05 & 92.51 & 74.41 & 84.88 \\
%Qwen3.5-122B-A10B & 39.50 & 55.97 & 61.80 & 62.85 & 55.89 & 74.73 & 55.12\\
Qwen3.5-122B-A10B & 64.95 & 87.80 & 84.34 & 90.45 & 92.82 & 75.02 & 84.37\\
%Qwen3.5-397B-A17B & 44.69 & 58.93 & 67.97 & 63.02 & 55.88 & \textbf{76.47} & 58.17\\
Qwen3.5-397B-A17B & \textbf{69.73} & \textbf{90.28} & \textbf{87.92} & 90.37 & 92.69 & \textbf{76.37} & \textbf{86.47}\\
%InternVL3.5-8B &  \textbf{45.83} & 62.27 & 70.69 & \textbf{65.60} & 57.03 & 57.79  & \textbf{60.46}\\
InternVL3.5-8B &  69.60 & 88.43 & 82.62 & \textbf{92.43} & 92.84 & 59.04  & 85.27\\
%InternVL3.5-14B &  44.80 & 61.85 & \textbf{71.56} & 65.26 & \textbf{57.22} & 61.19 & 60.33 \\
InternVL3.5-14B &  69.13 & 87.41 & 85.65 & 91.80 & \textbf{93.03} & 61.96 & 85.61 \\
%InternVL3.5-38B & 43.40 & \textbf{62.29} & 69.43 & 63.83 & 56.71 & 57.75 & 59.36\\
InternVL3.5-38B & 68.01 & 89.16 & 82.24 &90.95 & 92.64   & 59.71 & 84.85\\
% DeepSeek-V3.2-Speciale & 40.62 & 46.25 & 58.81 & 61.61 & 44.65 & 28.99 & 50.24\\
\hline
\end{tabular}
% \end{adjustbox}
\caption{DRFR scores (\%) of different constraint domains from evaluated MLLMs. The best score in each column is bolded separately for closed-source and open-source MLLMs. }
\label{tab:domains}
\end{table*}
}

%\textbf{Declarative Memory} represents the average scores across explicit memory tasks, \textbf{Non-declarative} represents average scores across implicit memory tasks, and \textbf{Overall} represents the average performance of models between declarative memory and non-declarative memory. 

\subsection{Scientific Answer Correctness Evaluation}
\label{app:answer_correctness}

We separately evaluate whether each model response correctly answers the underlying scientific problem. 
We utilize CompassVerifier-32B~\cite{liu2025compassverifier} to evaluate the answer correctness. The judge model is provided with the original question, the visual input when available, the reference answer, and the model response, and returns a binary correctness label. This evaluation focuses only on scientific correctness and does not penalize violations of injected constraints unless they alter the scientific meaning of the answer. The resulting labels are combined with instruction-adherence results to classify
responses as Correct Followed (CF), Correct Violated (CV), Incorrect Followed (IF), or Incorrect Violated (IV).

%%%%%%%%%%%%%%%%%%%%%%%%%%%%%%%%%%%%%%%%%%%%%%%%%%%%%%%%%%%%%%%%%%%

\subsection{Constraint Verification Protocols}
% 说明三种或多种评测方式：
We adopt 2 constraint verification protocols to judge the instruction-following capability of models.

\textbf{Script-Based Evaluation}
We employ programmatic verification when constraint compliance can be determined using explicit and reproducible criteria. This category includes numerical-format and casing checks, structured parsing of JSON, lists, tuples, dictionaries, HTML, XML, and CSV, and domain-specific validation with tools such as RDKit. Regular-expression matching is used to identify formulas, answer markers, option labels, scientific notation, and other predefined textual patterns. For constraints requiring semantic extraction, an LLM may extract a molecular representation or scientific method from the response. However, the extracted content is subsequently verified using regular expressions, structured parsers, or RDKit, and the LLM does not make the final compliance decision. The extractor model in the experiment is GPT-4.1.

\textbf{LLM-as-a-Judge}
We use an LLM as the evaluator when constraint satisfaction requires semantic interpretation that cannot be reliably determined through fixed rules. The judge compares the augmented question, constraint parameters, and model response to determine whether the required reasoning or reaction steps are covered and to calculate the proportion of matched steps. It is also used to determine whether a response contains exactly the required number of semantically distinct response categories. The judge is instructed to return a structured decision and justification, which are parsed into the final compliance score. We select GPT-4.1 as the judge model.

% Exact Match；
% Rule-based Verification；
% Precision-based Verification；
% Domain-specific Parser；
% LLM-as-a-Judge。

%%%%%%%%%%%%%%%%%%%%%%%%%%%%%%%%%%%%%%%%%%%%%%%%%%%%%%%%%%%%%%%%%%%
\section{Additional Results and Analysis}
\label{app:additional_results}

\subsection{Complete results over scientific and general constraints}

We illustrate complete results of models over 5 scientific constraint domains and general constraint domain in Table~\ref{tab:domains} to support the conclusion in the section Performance on Constraint Domains.

\subsection{Analysis of the Significance of Correctness and Instruction Following}
\label{app:significance}
\begin{table}[t]
\centering

\begin{threeparttable}
\begin{tabular}{
  p{0.16\textwidth}
  p{0.05\textwidth}
  p{0.05\textwidth} 
  p{0.12\textwidth} }
\toprule
Model&$\phi$ & Jaccard & $p$-value  \\
\midrule
GPT-5.2             &0.1178 & 0.3585  & $3.21 \times 10^{-9}$   \\
Grok-4-Fast         &0.0081 & 0.2930  & 0.68                     \\
Gemini-3.1-Pro-Preview &0.1548 &0.3538 & $7.35 \times 10^{-15}$  \\
Claude-Sonnet-4-6     &0.2069 & 0.3640 & $4.31 \times 10^{-25}$  \\
Qwen3.5-397B-A17B    &0.0672 & 0.3300 & $7.68 \times 10^{-4}$   \\
InternVL3.5-14B     & 0.1459 & 0.2704   & $2.33 \times 10^{-13}$  \\
\bottomrule
\end{tabular}
\end{threeparttable}
\caption{Within-model associations between scientific correctness and instruction following. $\phi$ coefficient measures the direction and strength of the binary association, Jaccard measures the overlap between the two outcomes, and the p-values are obtained from two-sided Pearson $\chi^2$ tests of independence.}
\label{tab:within-model-p-values}
\end{table}

To examine the within-model association between scientific correctness and instruction following, we construct a \(2\times2\) contingency table for each model. Let \(A\) denote scientific correctness and \(B\) denote instruction following. The four observed cell counts are \(CF\), \(CV\), \(IF\) , and \(IV\):

\[
\begin{array}{c|cc}
 & B=1 & B=0\\ \hline
A=1 & CF & CV\\
A=0 & IF & IV
\end{array}.
\]

We apply the two-sided Pearson \(\chi^2\) test of independence separately to each model. The null hypothesis is

\begin{equation}
    H_0:\ A\text{ and }B\text{ are independent}\quad(\phi=0),
\end{equation}

whereas the alternative hypothesis is

\begin{equation}
    H_1:\ A\text{ and }B\text{ are not independent}\quad(\phi\neq0).
\end{equation}

For each cell, the expected count under \(H_0\) is calculated as

\begin{equation}
    E_{ij}=\frac{(\text{row total}_i)(\text{column total}_j)}{n},
\end{equation}

and the Pearson test statistic is

\begin{equation}
    \chi^2=\sum_{i=1}^{2}\sum_{j=1}^{2}
\frac{(O_{ij}-E_{ij})^2}{E_{ij}},
\end{equation}

where \(O_{ij}\) and \(E_{ij}\) are the observed and expected counts, respectively. Because a \(2\times2\) table has one degree of freedom, the two-sided \(p\)-value is obtained from the upper tail of a \(\chi^2_1\) distribution:

\begin{equation}
    p=\Pr\!\left(\chi^2_1\geq\chi^2_{\mathrm{obs}}\mid H_0\right)
  =1-F_{\chi^2_1}\!\left(\chi^2_{\mathrm{obs}}\right).
\end{equation}

Thus, a small \(p\)-value indicates that the observed discrepancy from independence would be unlikely under \(H_0\). We reject \(H_0\) when \(p<0.05\). Because six model-specific tests are conducted, we additionally apply a Bonferroni-corrected threshold of

\begin{equation}
    \alpha^{*}=\frac{0.05}{6}=0.00833.
\end{equation}

The direction and magnitude of association are quantified using the \(\phi\) coefficient:

\begin{equation}
    \phi=
\frac{CF\cdot IV-CV\cdot IF}
{\sqrt{(CF+CV)(IF+IV)(CF+IF)(CV+IV)}}.
\end{equation}

Equivalently, for a \(2\times2\) table,

\begin{equation}
    |\phi|=\sqrt{\frac{\chi^2}{n}},
\end{equation}

with its sign determined by \(CF\cdot IV-CV\cdot IF\). The coefficient ranges from \(-1\) to \(1\). \(\phi>0\) indicates that correctness and instruction following tend to occur together, \(\phi<0\) indicates an inverse association, and values close to zero indicate weak association.

As a complementary descriptive measure, we calculate the Jaccard coefficient:

\begin{equation}
    J=\frac{CF}{CF+CV+IF}.
\end{equation}

It represents the proportion of responses satisfying both criteria among those satisfying at least one. The \(IV\) cell is excluded because those responses satisfy neither criterion. %Unlike the Pearson test, the Jaccard coefficient is not associated with a default null hypothesis or \(p\)-value; formal between-model comparisons would require confidence intervals and an additional procedure such as bootstrap or permutation testing.

%Table~\ref{tab:within-model-p-values} demonstrates that scientific correctness and instruction following are statistically related but remain distinct and only weakly coupled capabilities. As reported in Table~\ref{tab:within-model-p-values}, five of the six models exhibit a statistically significant positive association under the two-sided Pearson \(\chi^2\) test of independence. All five results remain significant after Bonferroni correction, indicating that the conclusions are robust to multiple testing rather than artifacts of repeated comparisons.

%Statistical significance, however, does not imply strong practical coupling. Figure~\ref{fig:phi} shows that the significant associations have uniformly small positive \(\phi\) coefficients, while Grok-4-fast exhibits virtually no association. The similarly modest Jaccard coefficients show that only \(27.04\%\)–\(36.40\%\) of responses satisfying at least one criterion satisfy both simultaneously. Taken together, the \(p\)-values quantify the evidence against independence, the \(\phi\) coefficients characterize the direction and limited strength of the association, and the Jaccard coefficients describe the modest overlap between the two desirable outcomes.

%These findings expose a systematic capability gap: instruction adherence is not a reliable proxy for scientific correctness, and scientifically correct answers do not necessarily satisfy the specified constraints. Consequently, optimizing either capability in isolation is insufficient; models should be explicitly trained and evaluated for their joint attainment.

 Table~\ref{tab:within-model-p-values} demonstrates that scientific correctness and instruction following are statistically related but remain distinct and only weakly coupled capabilities. Under the two-sided Pearson $\chi^2$ test of independence, five of the six models exhibit statistically significant positive associations. These five results remain significant after Bonferroni correction (\(\alpha^{*}=0.00833\)), indicating that the conclusions are robust to multiple testing rather than artifacts of repeated comparisons. Grok-4-fast is the only exception, with \(p=0.68\) and a near-zero \(\phi\) coefficient (\(\phi=0.0081\)), providing no significant evidence against independence.

Statistical significance, however, does not imply strong practical coupling. As shown in Table~\ref{tab:within-model-p-values}, the \(\phi\) coefficients are modest even for the statistically significant models, ranging from \(0.0672\) to \(0.2069\). The Jaccard coefficients show that only \(27.04\%\)–\(36.40\%\) of responses satisfying at least one criterion satisfy both simultaneously. Taken together, the \(p\)-values quantify the evidence against independence, the positive but modest \(\phi\) coefficients characterize the direction and limited strength of the associations, and the Jaccard coefficients describe the limited overlap between the two desirable outcomes.

These findings expose a systematic capability gap that instruction adherence is not a reliable proxy for scientific correctness, and scientifically correct answers do not necessarily satisfy the specified constraints. Consequently, optimizing either capability in isolation is insufficient, and models should be explicitly trained and evaluated for their joint attainment.

\subsection{Results across Source Datasets}
To determine whether discipline-level averages obscure variation among the heterogeneous source datasets, we further disaggregate GPT-5.2's performance within Material. This setting holds the model and discipline fixed while comparing all four material sources in SciMIF. As shown in Table~\ref{tab:per_source_material}, DRFR varies substantially across sources, ranging from 63.92\% on LLM4Mat-Bench to 96.28\% on MatSciBench. In particular, LLM4Mat-Bench is 22.55 percentage points below the next-lowest source, MatCha, and 32.36 points below MatSciBench. This result shows that the aggregate materials science score does not imply uniform instruction-following ability across its constituent datasets. 
{
\begin{table}[t]
\centering
\small
\resizebox{\linewidth}{!}{%
\begin{tabular}{l|cccc}
\hline
& \textbf{MaScQA}
& \textbf{Matcha}
& \textbf{LLM4Mat-Bench}
& \textbf{MatSciBench}
\\
\hline
DRFR & 93.19 & 86.47 & 63.92 & 96.28
\\
\hline
\end{tabular}%
}
\caption{Per-source DRFR (\%) of GPT-5.2 on materials science.}
\label{tab:per_source_material}
\end{table}
}

\subsection{Text-Only and Multimodal Performance}
%review L16, rebuttal line 95-99
To examine modality-associated differences in instruction-following performance, we restrict the analysis to Geography, Material, and Physics, the three disciplines in SciMIF that contain both text-only and multimodal samples. Chemistry and Biology are excluded because their samples
are exclusively text-based. We use Gemini-3.1-Pro-Preview as a representative model and compare the two modalities within each discipline, thereby holding the evaluated model fixed and reducing cross-disciplinary confounding. As shown in Table~\ref{tab:modality_results}, the multimodal subsets consistently obtain lower DRFR scores than their text-only counterparts: 64.67\% versus 87.92\% in Geography, 85.50\% versus 90.16\% in Material, and 74.49\% versus 83.76\% in Physics. These values correspond to absolute decreases of
23.25, 4.66, and 9.27 percentage points, respectively. The consistently lower scores provide evidence of a modality-associated degradation in instruction compliance, while the heterogeneous gaps  indicate that this pattern varies across disciplinary contexts.

{
\begin{table}[t]
\centering
\small

\begin{tabular}{l|ccc}
\hline
\textbf{Modality}
& \textbf{Geography}
& \textbf{Material}
& \textbf{Physics}
\\
\hline
Multimodal & 64.67 & 85.50 & 74.49 \\
Text-only & 87.92 & 90.16 & 83.76 \\
\hline
\end{tabular}
\caption{DRFR scores (\%) of Gemini-3.1-Pro-Preview on multimodal and
text-only subsets.}
\label{tab:modality_results}
\end{table}
}

\subsection{Correctness Contrast}
%一个模型一个柱状图，加入指令题目更复杂

\begin{figure*}[t]
    \centering

    \begin{subfigure}[t]{0.49\textwidth}
        \centering
        \includegraphics[width=\linewidth]{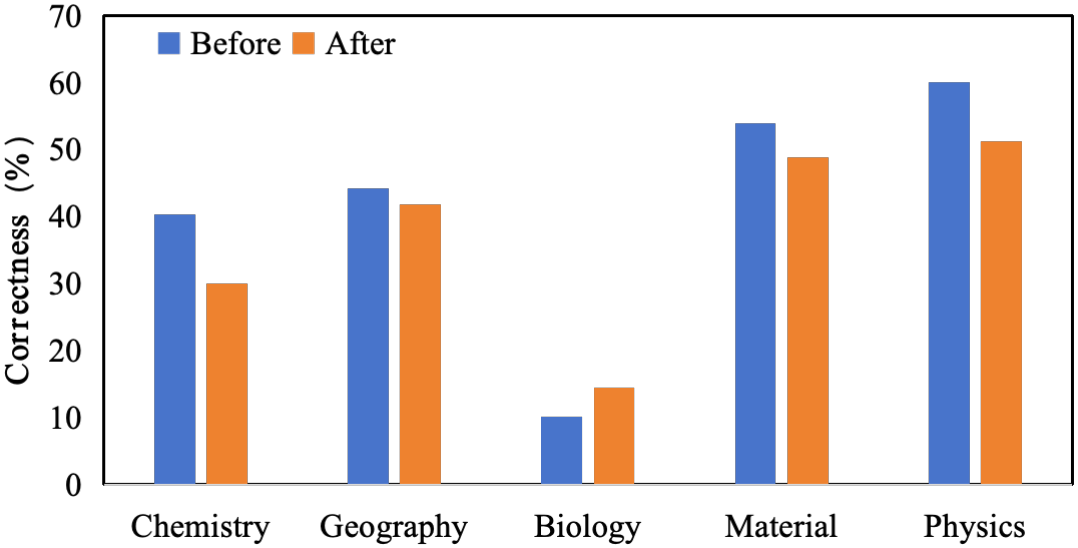}
        \caption{GPT-5.2.}
        \label{fig:correctness_gpt}
    \end{subfigure}
    \hfill
    \begin{subfigure}[t]{0.49\textwidth}
        \centering
        \includegraphics[width=\linewidth]{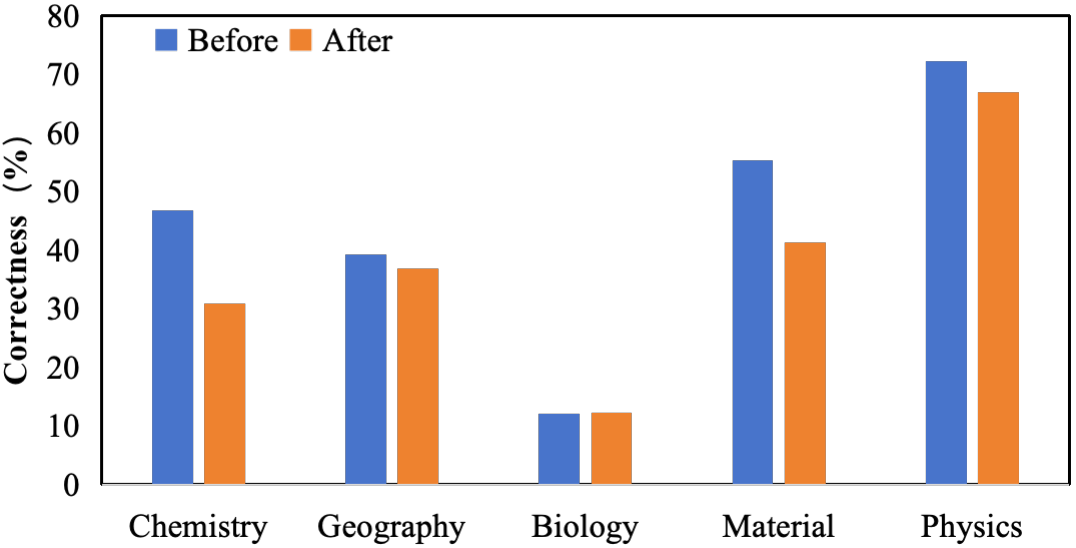}
        \caption{Qwen3.5-397B-A17B.}
        \label{fig:correctness_qwen}
    \end{subfigure}

    \caption{
    Scientific answer correctness before and after constraint
    augmentation for two representative models across the five
    disciplines. \textit{Before} denotes evaluation on the original
    scientific questions, while \textit{After} denotes evaluation
    after compatible scientific and general constraints are added.
    The comparison shows how additional instruction-following
    requirements are associated with changes in scientific answer
    correctness.
    }
    \label{fig:correctness_comparison}
\end{figure*}

To investigate how instructions influence task difficulty, we evaluated GPT-5.2 and Qwen3.5-397B-A17B on the original and constraint-augmented queries. The results in Figure~\ref{fig:correctness_comparison} indicate that, in the absence of explicit instructions, the models generally achieve higher accuracy across most domains. However, when tasks include additional requirements, the models’ attention to instructions incurs an accuracy penalty, suggesting a potential degradation of capability in settings that demand strict output adherence, such as agent pipelines. This trade-off highlights the challenge of simultaneously maintaining model reliability and usability, and motivates further research into methods that enhance both scientific rigor and practical effectiveness.

% To investigate the impact of instructions on task difficulty, we select GPT-5.2 and Qwen3.5-397B-A17B and conducted a comparative analysis of their performance with and without constraints. The experimental results indicate that, under conditions where no instructions are provided, the models demonstrate a significantly higher rate of correct responses and exhibit greater robustness in their performance. This observation suggests that the introduced constraints may have increased the complexity of the task, thereby leading to a decline in the models' accuracy in scenarios where constraints are present. 

\subsection{Constraint-Level Error Analysis}
\begin{figure*}[!t]
  \centering 
  \includegraphics[width=\linewidth]{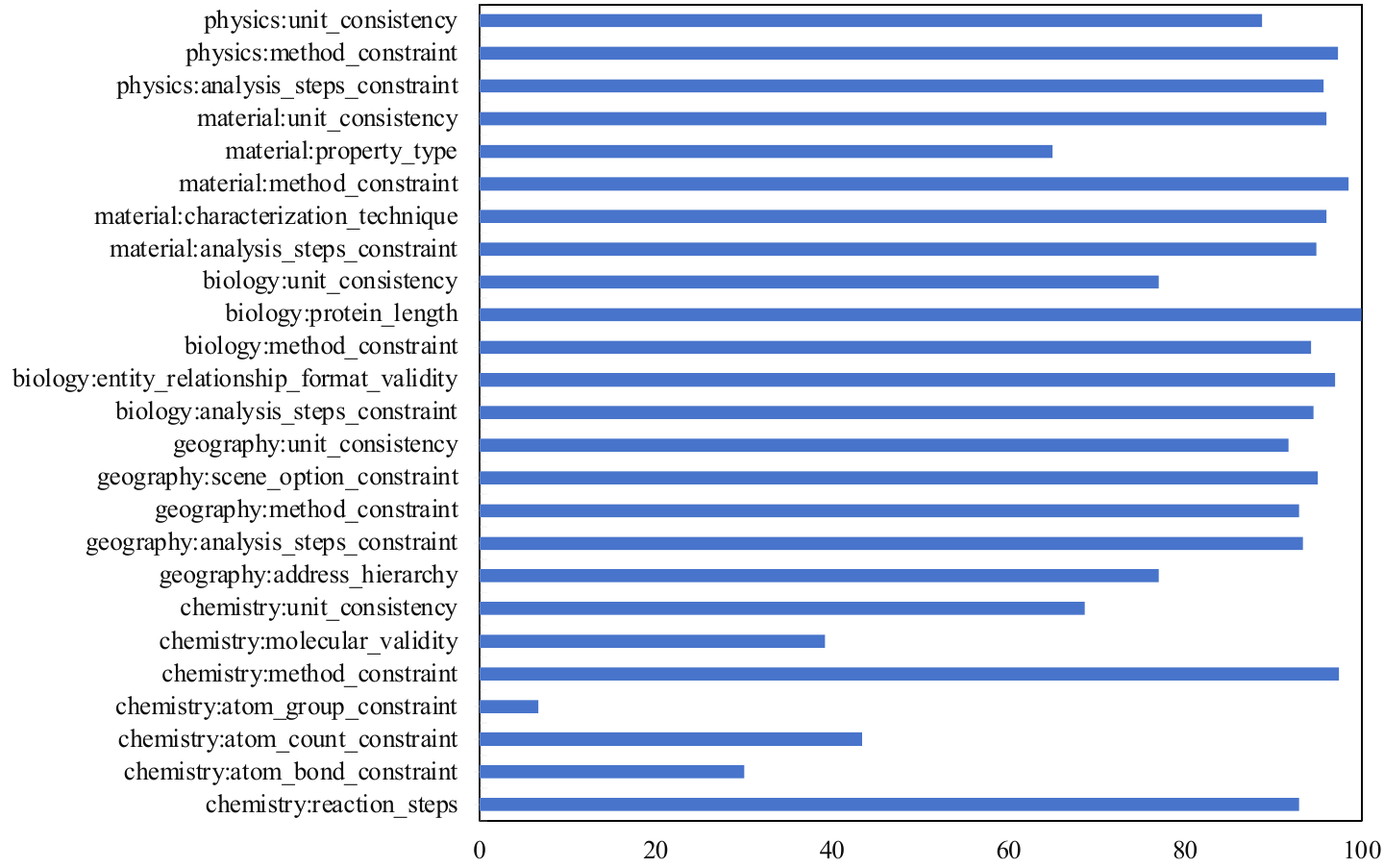}
\caption{GPT-5.2 performance across individual scientific constraints.}
  \label{fig:constraint performance}
\end{figure*}

% 缩写一段话即可 只需要介绍这个case不需要分析
To better understand the underlying mechanisms behind the performance disparities across different constraint types, we conduct a detailed analysis on GPT-5.2, whose DRFR scores for each constraints are shown in Figure ~\ref{fig:constraint performance}. The observations clearly indicate that GPT-5.2’s performance in executing chemistry constraints is significantly lower than its performance in other disciplines. This phenomenon aligns with the conclusion that GPT-5.2 achieved its lowest score in the field of chemistry, and further corroborates its areas of weakness within the subject. 

\begin{figure*}[!h]
  \centering 
  \includegraphics[width=\linewidth]{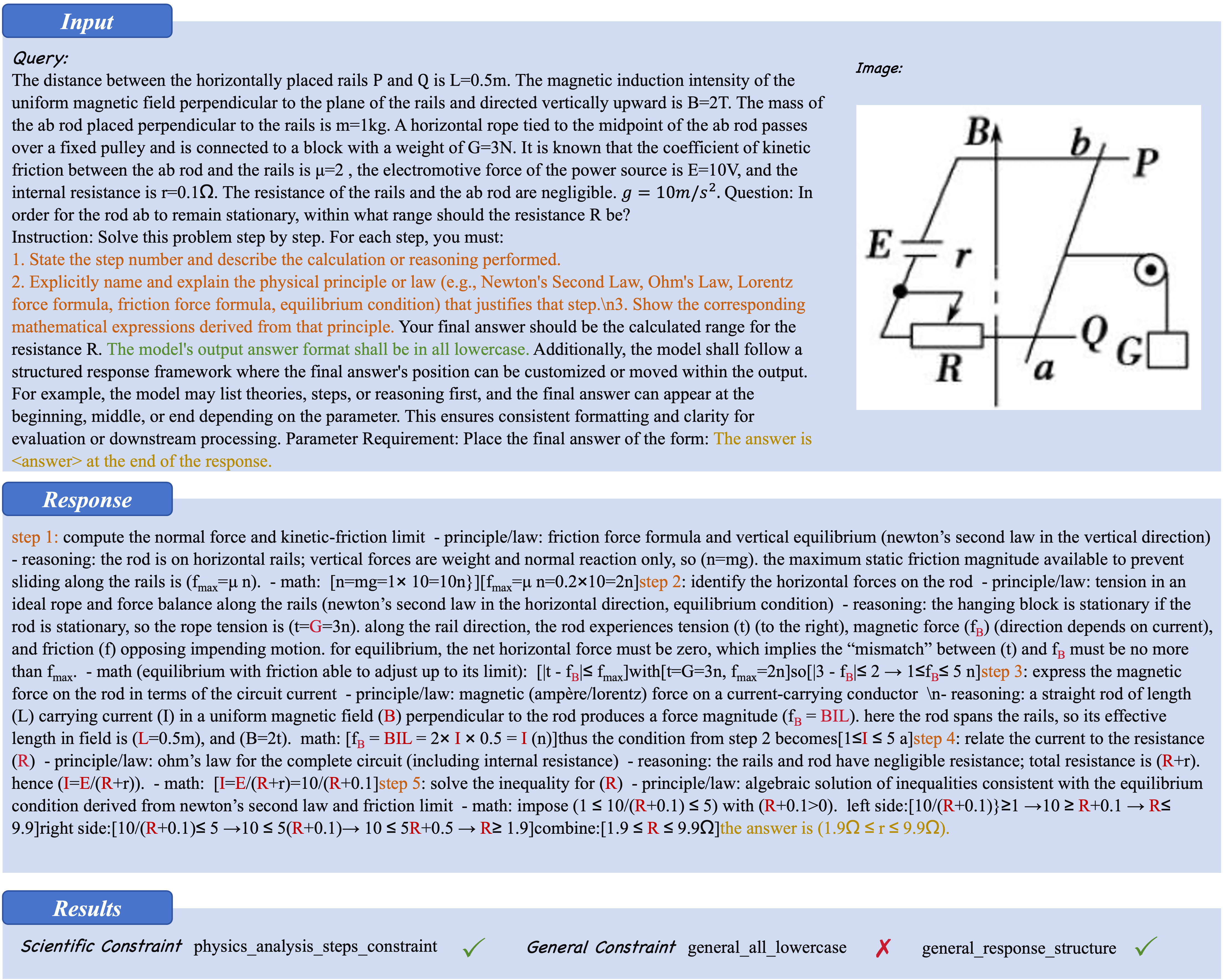}
\caption{A representative physics case illustrating partial constraint satisfaction. The response satisfies the required analysis procedure and response structure but violates the lowercase constraint.}
  \label{fig:case}
\end{figure*}

\subsection{Representative Case Studies}
\label{app:case_studies}
To provide a more concrete view of model behavior beyond aggregate scores, we analyze a representative physics example containing both scientific and general constraints. As shown in Figure~\ref{fig:case}, the instruction asks the model to solve a mechanics and electromagnetism problem while simultaneously satisfying three independently evaluated constraints. Specifically, the model is required to follow a step-by-step physics analysis procedure, organize the response according to a prescribed structure, and produce the entire output in lowercase.

The response correctly derives the maximum static-friction force, allowable magnetic force, current requirement, and circuit-resistance range using relevant physical principles. It therefore satisfies the  constraint named
\texttt{physics:analysis\_steps}. The final result,
$1.9\,\Omega \leq r \leq 9.9\,\Omega$, is also placed after the reasoning process, satisfying the
\texttt{general:response\_structure} constraint.

However, the response violates the constraint, which
is \texttt{general:all\_lowercase}  because uppercase symbols
such as $R$, $E$, $B$, and $L$ appear in the output. This case shows that correct scientific reasoning and structural compliance do not necessarily guarantee adherence to fine-grained symbolic requirements. It also demonstrates the value of constraint-level evaluation, which can identify the specific source of failure rather than treating the entire instruction as uniformly unsuccessful.

%%%%%%%%%%%%%%%%%%%%%%%%%%%%%%%%%%%%%%%%%%%%%%%%%%%%%%%%%%%%%%%%%%%

\section{Data Compliance and Release}
\label{app:data_compliance}

SciMIF is constructed by augmenting samples from 13 existing scientific datasets. All source datasets are used exclusively for academic research and are processed in accordance with their respective licenses, terms of use, and redistribution requirements. The original scientific
questions, visual inputs, and reference answers retain their original provenance, and the corresponding source datasets are cited in the main paper and documented in the Appendix.

Our construction process does not alter the underlying scientific problem or its reference answer. Instead, it adds compatible scientific and general constraints to the original query and records the injected constraints as independently evaluable metadata. Each released sample
will therefore include its augmented query, reference answer, constraint list, task type, discipline, and source-dataset identifier, subject to the redistribution permissions of the corresponding source dataset.

For source datasets that permit redistribution, the processed samples will be included directly in the SciMIF release while preserving the required attribution and license information. For datasets or visual assets whose licenses do not permit direct redistribution, we will not
republish the restricted content. Instead, we will provide source identifiers, data-processing scripts, and reconstruction instructions that allow eligible users to obtain the original data from its official source and reproduce the corresponding SciMIF samples.

The public release will include the constraint taxonomy, augmented queries, constraint annotations, evaluation configurations, deterministic verification scripts, LLM-as-a-Judge prompts, and model evaluation code. We will additionally provide documentation describing
the provenance and applicable license of each source dataset. This release strategy is intended to support reproducibility while respecting the ownership, attribution, and redistribution conditions of the original scientific resources.

%%%%%%%%%%%%%%%%%%%%%%%%%%%%%%%%%%%%%%%%%%%%%%%%%%%%%%%%%%%%%%%%%%%

% \section{Manual Verification}
% %review L19, L20, rebuttal line 100-106, line 111-114
% \noindent\textbf{Human agreement.}
% Manual verification assesses whether the injected requirements are expressed
% coherently and faithfully in each augmented query. Because this process focuses
% on linguistic and logical correspondence rather than re-solving the underlying
% scientific question, it requires limited specialized domain knowledge. Before
% full-scale verification, two annotators independently examined the same 20
% randomly sampled instances and agreed on all cases, yielding an observed
% agreement of 100\%.

% \noindent\textbf{Judge reliability.}
% %待添加引用
% For requirements involving semantic judgment, model outputs are evaluated
% using an LLM-as-a-judge. We adopt the judge model and prompting protocol used in established LLM evaluation studies~\cite{NEURIPS2023_91f18a12,followbench} and apply the same protocol throughout the evaluation.

% \noindent\textbf{Data compliance and openness.}
% All source datasets are used for academic research in accordance with their
% applicable licensing requirements. The complete SciMIF data and evaluation code
% will be publicly released to support reproducibility.
% %多模态的，长一点的都可以


\end{document}